\documentclass[10pt,twocolumn,letterpaper]{article}

\usepackage[pagenumbers]{cvpr} 

\newcommand\extrafootertext[1]{%
    \bgroup
    \renewcommand\thefootnote{\fnsymbol{footnote}}%
    \renewcommand\thempfootnote{\fnsymbol{mpfootnote}}%
    \footnotetext[0]{#1}%
    \egroup
}

\usepackage{graphicx}
\usepackage{amssymb}
\usepackage{booktabs}
\usepackage{xcolor}         
\usepackage{multirow}
\usepackage{color, colortbl}
\def\etal{\emph{et al.}\xspace}
\definecolor{Gray}{gray}{0.90}
\usepackage{times}
\usepackage{epsfig}
\usepackage{array}

\usepackage{dblfloatfix}
\usepackage{amsmath}
\usepackage{bm}
\usepackage{breqn}
\usepackage{wrapfig}
\usepackage{adjustbox}
\usepackage[rightcaption]{sidecap}

\usepackage[pagebackref=true,breaklinks=true,colorlinks,bookmarks=false,citecolor=blue,linkcolor=blue,urlcolor=blue]{hyperref}

\usepackage{amssymb}
\usepackage{pifont}
\usepackage[capitalize]{cleveref}
\crefname{section}{Sec.}{Secs.}
\Crefname{section}{Section}{Sections}
\Crefname{table}{Table}{Tables}
\crefname{table}{Tab.}{Tabs.}

\newcommand{\txt}[1]{{\texttt{#1}}}
\usepackage[misc]{ifsym}
\newcommand{\corrAuthor}{$^{\textrm{\Letter}}$}
\usepackage[hang,flushmargin]{footmisc}

\def\cvprPaperID{} 
\def\confName{CVPR}
\def\confYear{2023}

\begin{document}

\title{UNETR++: Delving into Efficient and Accurate 3D Medical Image Segmentation}

\author{%
  Abdelrahman Shaker$^{1}$\corrAuthor \quad 
  Muhammad Maaz$^{1}$ \quad 
  Hanoona Rasheed$^{1}$ \quad
  Salman Khan$^{1,2}$ \\
  Ming-Hsuan Yang$^{3,4,5}$ \quad
  Fahad Shahbaz Khan$^{1,6}$
  \vspace{0.5em} \\
  $^{1}$Mohamed bin Zayed University of AI \quad 
  $^{2}$Australian National University \\
  $^{3}$University of California, Merced \quad
  $^{4}$Yonsei University \quad
  $^{5}$Google Research \quad
  $^{6}$Link\"{o}ping University 
  }
\maketitle
\extrafootertext{
\textsuperscript{\corrAuthor}\txt{abdelrahman.youssief@mbzuai.ac.ae}}
\begin{abstract}\vspace{-1em}
Owing to the success of transformer models, recent works study their applicability in 3D medical segmentation tasks. Within the transformer models, the self-attention mechanism is one of the main building blocks that strives to capture long-range dependencies. However, the self-attention operation has quadratic complexity which proves to be a computational bottleneck, especially in volumetric medical imaging, where the inputs are 3D with numerous slices.  In this paper, we propose a 3D medical image segmentation approach, named UNETR++, that offers both high-quality segmentation masks as well as efficiency in terms of parameters, compute cost, and inference speed. The core of our design is the introduction of a novel efficient paired attention (EPA) block that efficiently learns spatial and channel-wise discriminative features using a pair of inter-dependent branches based on spatial and channel attention.
Our spatial attention formulation is efficient having linear complexity with respect to the input sequence length. To enable communication between spatial and channel-focused branches, we share the weights of query and key mapping functions that provide a complimentary benefit (paired attention), while also reducing the overall network parameters. Our extensive evaluations on five benchmarks, Synapse, BTCV, ACDC, BRaTs, and Decathlon-Lung, reveal the effectiveness of our contributions in terms of both efficiency and accuracy. On Synapse, our UNETR++ sets a new state-of-the-art with a Dice Score of 87.2\%, while being significantly efficient with a reduction of over 71\% in terms of both parameters and FLOPs, compared to the best method in the literature. Code: \url{https://tinyurl.com/2p87x5xn}.
\end{abstract}

\section{Introduction}

\begin{figure*}[t]
  \centering
    \includegraphics[width=0.96\linewidth]{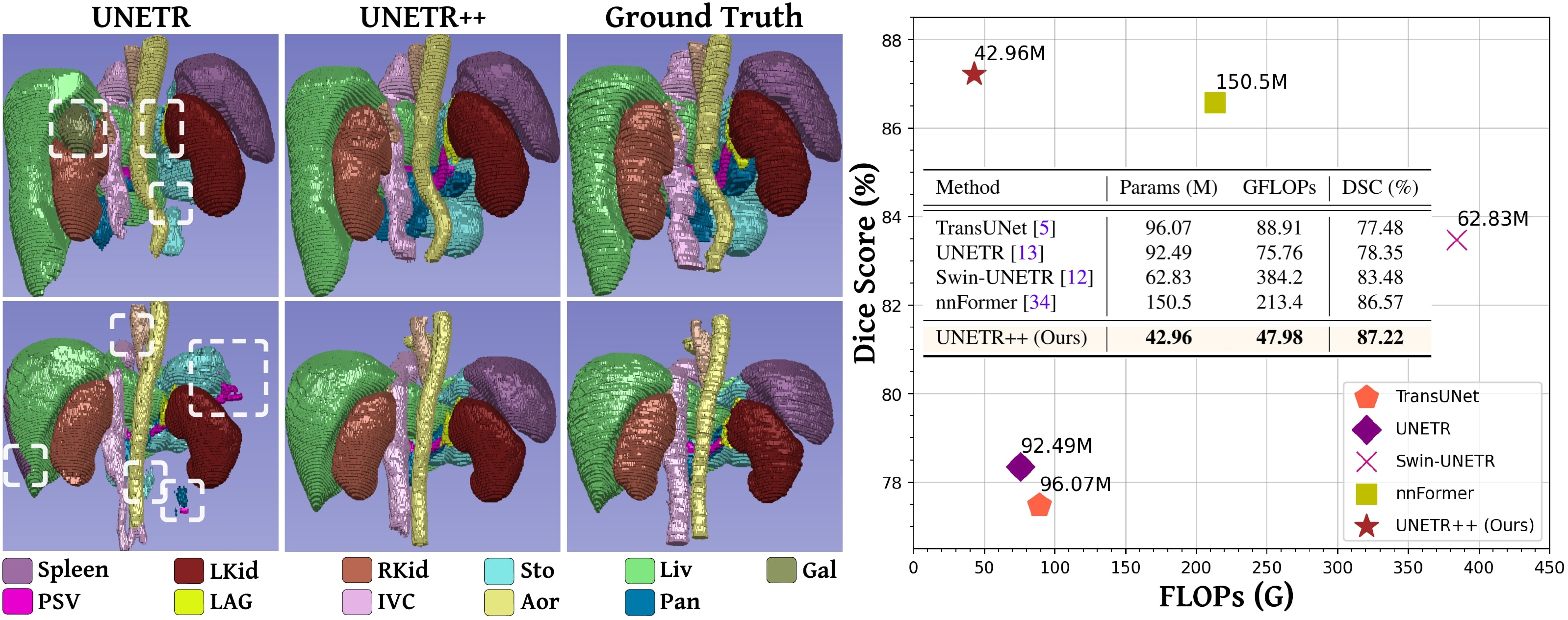}
    \vspace{-0.45em} 
    \caption{\textbf{Left:} Qualitative comparison between the baseline UNETR~\cite{UNETR} and our UNETR++ on Synapse. We present two examples containing multiple organs. Each inaccurate segmented region is marked with a white dashed box.  
    In the first row, UNETR struggles to accurately segment the \textit{right kidney} (RKid) and confuses it with \textit{gallbladder} (Gal). Further, both the \textit{stomach} (Sto) and \textit{left adrenal gland} (LAG) tissues are inaccurately segmented. In the second row, UNETR struggles to segment the whole \textit{spleen} and mixes it with \textit{stomach} (Sto) and \textit{portal and splenic veins} (PSV). Moreover, it under and over-segments certain organs (\eg, PSV and Sto). In comparison, our UNETR++ that efficiently encodes enriched inter-dependent spatial and channel features within the proposed EPA block, accurately segments all organs in these examples. Best viewed zoomed in. Additional qualitative comparisons are presented in Fig.~\ref{Fig:Qualitative_results} and supplementary material. \textbf{Right:} Accuracy (Dice score) \vs model complexity (FLOPs and parameters) comparison on Synapse. Compared to best existing nnFormer~\cite{nnFormer}, UNETR++ achieves better segmentation performance while significantly reduces the model complexity by over 71\%.}
    \label{fig:Intro}
    \vspace{-0.5cm}
\end{figure*}

\label{sec:intro}
\noindent Volumetric (3D) segmentation is a fundamental problem in medical imaging with numerous applications including, tumor identification and organ localization for diagnostic purposes~\cite{UNETR, nnUNet}. The task is typically addressed by utilizing a U-Net~\cite{UNet} like encoder-decoder architecture where the encoder generates a hierarchical low-dimensional representation of a 3D image and the decoder maps this learned representation to a voxel-wise segmentation. Earlier CNN-based methods use convolutions and deconvolutions in the encoder and the decoder, respectively, but struggle to achieve accurate segmentation results likely due to their limited receptive field. In contrast, transformer-based methods are inherently global and have recently demonstrated competitive performance at the cost of increased model complexity.

Recently, several works~\cite{UNETR,nnFormer,SWIN_UNETR} have explored designing hybrid architectures to combine the merits of both local convolutions and global attention. 
While some approaches~\cite{UNETR} use transformer-based encoder with convolutional decoder, others~\cite{nnFormer,SWIN_UNETR} aim at designing hybrid blocks for both encoder and decoder subnetworks. However, these works mainly focus on increasing the segmentation accuracy which in turn substantially increases the model sizes in terms of both parameters and FLOPs, leading to unsatisfactory robustness. We argue that this unsatisfactory robustness is likely due to their inefficient self-attention design, which becomes even more problematic in volumetric medical image segmentation tasks. Further, these existing approaches do not capture the explicit dependency between spatial and channel features which can improve the segmentation quality. In this work, we aim to simultaneously improve both the segmentation accuracy \textit{and} the model efficiency in a single unified framework.  

\noindent\textbf{Contributions:} We propose an efficient hybrid hierarchical architecture for 3D medical image segmentation, named UNETR++, that strives to achieve both better segmentation accuracy and efficiency in terms of parameters, FLOPs, and inference speed. Built on the recent UNETR framework~\cite{UNETR}, our proposed UNETR++ hierarchical approach introduces a novel  
\textit{efficient paired attention} (EPA) block that efficiently captures enriched inter-dependent spatial and channel features by applying both spatial and channel attention in two branches. Our spatial attention in EPA projects the keys and values to a fixed lower dimensional space, making the self-attention computation linear with respect to the number of input tokens. On the other hand, our channel attention emphasizes the dependencies between the channel feature maps by performing the dot-product operation between queries and keys in the channel dimension. Further, to capture a strong correlation between the spatial and channel features, the weights for queries and keys are shared across the branches which also aids in controlling the number of network parameters. In contrast, the weights for values are kept independent to enforce learning complementary features in both branches. 

We validate our UNETR++ approach by conducting comprehensive experiments on five benchmarks: Synapse~\cite{BTCV}, BTCV~\cite{BTCV}, ACDC~\cite{ACDC}, BRaTs~\cite{BRATS}, and Decathlon-Lungs~\cite{MSD}. Both qualitative and quantitative results demonstrate the effectiveness of UNETR++, leading to better performance in terms of segmentation accuracy \textit{and} model efficiency compared to the existing methods in the literature.
On Synapse, UNETR++ achieves high-quality segmentation masks (see Fig.\ref{fig:Intro} left) with an absolute gain of 8.9\% in terms of Dice Score while significantly reducing the model complexity with a reduction of 54\% in terms of parameters and 37\% in FLOPs, compared to the baseline UNETR~\cite{UNETR}. Further, UNETR++ outperforms the best existing nnFormer~\cite{nnFormer} method with a considerable reduction in terms of both parameters and FLOPs (see Fig.\ref{fig:Intro} right).

\section{Related Work}
\noindent\textbf{CNN-based Segmentation Methods}: 
Since the introduction of the U-Net design~\cite{UNet}, several CNN-based approaches~\cite{zhu2017deeply, zhou2018unet++, huang2020unet, cai2020dense} have extended the standard U-Net architecture for various medical image segmentation tasks. In the case of 3D medical image segmentation~\cite{gibson2018automatic, dou20163d, cciccek20163d, milletari2016v, valanarasu2021medical}, the full volumetric image is typically processed as a sequence of 2D slices. Several works have explored hierarchical frameworks to capture contextual information. Milletari \etal{\cite{milletari2016v}} propose to use 3D representations of the volumetric image by down-sampling the volume to lower resolutions for preserving the beneficial image features. Çiçek \etal{\cite{cciccek20163d}} extend the U-Net architecture to volumetric segmentation by replacing the 2D operations with their 3D counterparts, learning from sparsely annotated volumetric images. Isensee \etal{\cite{nnUNet}} introduce a generalized segmentation framework, named nnUNet, that automatically configures the architecture to extract features at multiple scales. Roth \etal{\cite{roth2017hierarchical}} propose a multi-scale 3D fully convolution network to learn representations from varying resolutions for multi-organ segmentation. Further, several efforts in the literature have been made to encode holistic contextual information within CNN-based frameworks using, \eg, image pyramids ~\cite{zhao2017pyramid}, large kernels ~\cite{peng2017large}, dilated convolution ~\cite{chen2018encoder}, and deformable convolution ~\cite{li2020pgd}. 

\noindent\textbf{Transformers-based Segmentation Methods}: 
Vision transformers (ViTs) have recently gained popularity thanks to their ability to encode long-range dependencies leading to promising results on various vision tasks, including classification~\cite{ViTs} and detection~\cite{DETR}. One of the main building blocks within the transformer's architecture is the self-attention operation that models the interactions among the sequence of image patches, thereby learning global relationships. Few recent works have explored alleviating the complexity issue of standard self-attention operation within transformer frameworks~\cite{SparseAttention, LinFormer, Maaz2022EdgeNeXt, Reformer}. However, most of these recent works mainly focus on the classification problem and have not been studied for dense prediction tasks. \\
In the context of medical image segmentation, few recent works ~\cite{karimi2021convolution, cao2021swin} have investigated pure transformers designs. Karimi \etal{\cite{karimi2021convolution}} propose to divide a volumetric image into 3D patches which are then flattened to construct a 1D embedding and passed to a backbone for global representations. Cao \etal{\cite{cao2021swin}} introduce an architecture with shifted windows for 2D medical image segmentation. Here, an image is divided into patches and fed into a U-shaped encoder-decoder for local-global representation learning.  

\noindent\textbf{Hybrid Segmentation Methods}: Other than pure CNN or transformers-based designs, several recent works~\cite{TransFuse, valanarasu2021medical, TransUNet, lin2022ds, UNETR, nnFormer} have explored hybrid architectures to combine convolution and self-attention operations for better segmentation. TransFuse~\cite{TransFuse} proposes a parallel CNN-transformer architecture with a BiFusion module to fuse multi-level features in the encoder. MedT~\cite{valanarasu2021medical} introduces a gated position-sensitive axial-attention mechanism in self-attention to control the positional embedding information in the encoder, while the ConvNet module in the decoder produces a segmentation model. TransUNet ~\cite{TransUNet} combines transformers and the U-Net architecture, where transformers encode the embedded image patches from convolution features and the decoder combines the upsampled encoded features with high-resolution CNN features for localization. Ds-transunet~\cite{lin2022ds} utilizes a dual-scale encoder based on Swin transformer~\cite{Swin} to handle multi-scale inputs and encode local and global feature representations from different semantic scales through self-attention. Hatamizadeh \etal{\cite{UNETR}} introduce a 3D hybrid model, UNETR, that combines the long-range spatial dependencies of transformers with the CNN's inductive bias into a “U-shaped” encoder-decoder architecture. The transformer blocks in UNETR are mainly used in the encoder to extract fixed global representations and then are merged at multiple resolutions with a CNN-based decoder. Zhou \etal{\cite{nnFormer}} introduce an approach, named nnFormer, that adapts the Swin-UNet~\cite{cao2021swin} architecture. Here,  
convolution layers transform the input scans into 3D patches and volume-based self-attention modules are introduced to build hierarchical feature pyramids. While achieving promising performance, the computational complexity of nnFormer is significantly higher compared to UNETR and other hybrid methods.

\noindent\textbf{Our Approach}: As discussed above, most recent hybrid approaches, such as UNETR~\cite{UNETR} and nnFormer~\cite{nnFormer}, achieve improved segmentation performance compared to their pure CNNs and transformers-based counterparts. However, we note that this pursuit of increasing the segmentation accuracy by these hybrid approaches comes at the cost of substantially larger models (both in terms of parameters and FLOPs), which can further lead to unsatisfactory robustness.
For instance, UNETR achieves favorable accuracy but comprises 2.5$\times$ more parameters, compared to the best existing CNN-based nnUNet~\cite{nnUNet}. Moreover, nnFormer obtains improved performance over UNETR but further increases the parameters by 1.6$\times$ and FLOPs by 2.8$\times$. Furthermore, we argue that these aforementioned hybrid approaches struggle to effectively capture the inter-dependencies between feature channels to obtain an enriched feature representation that encodes both the spatial information as well as the inter-channel feature dependencies. In this work, we set out to collectively address the above issues in a unified hybrid segmentation framework.

\section{Method}
\begin{figure*}[t]
  \centering
    \includegraphics[width=1.0\linewidth]{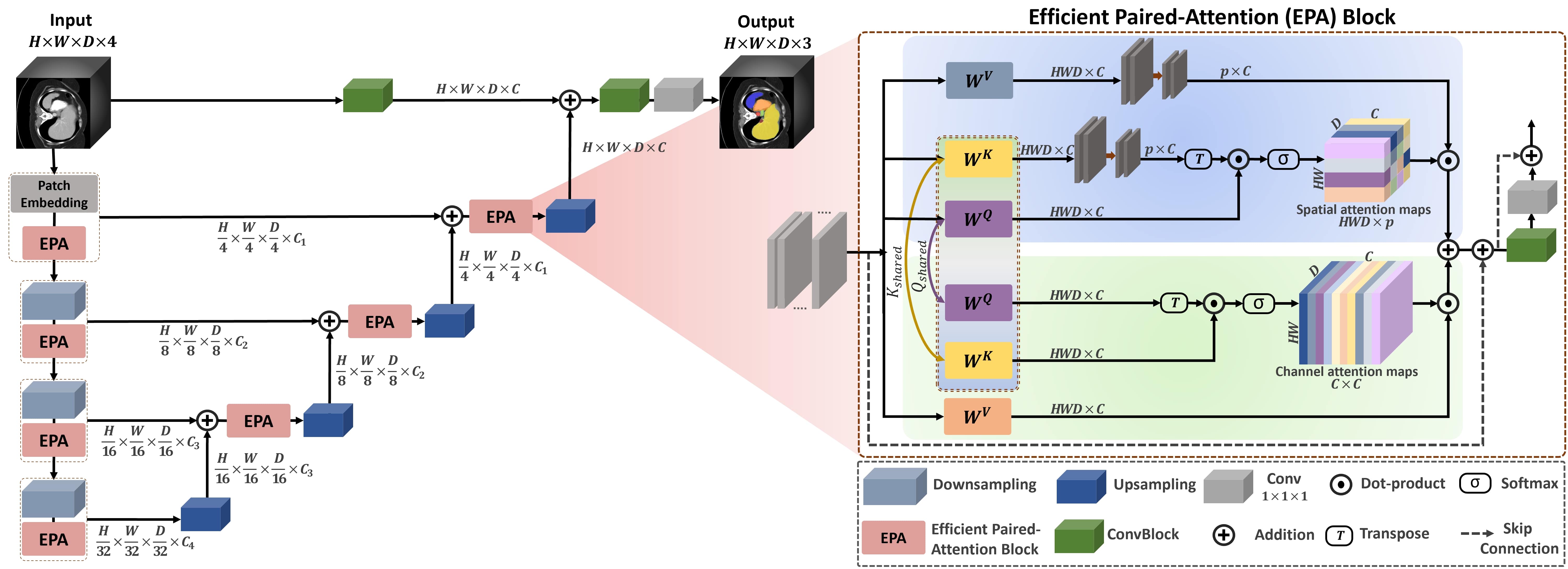}
    \vspace{-1.2em} 
    \caption{Overview of our UNETR++ approach with hierarchical encoder-decoder structure. The 3D patches are fed to the encoder, whose outputs are then connected to the decoder via skip connections followed by convolutional blocks to produce the final segmentation mask. The focus of our design is the introduction of an \textit{efficient paired-attention} (EPA) block (Sec.~\ref{sec:EPA}). Each EPA block performs two tasks using parallel attention modules with shared keys-queries and different value layers to efficiently learn enriched spatial-channel feature representations. As illustrated in the EPA block diagram (on the right), the first (top) attention module aggregates the spatial features by a weighted sum of the projected features in a linear manner to compute the spatial attention maps, while the second (bottom) attention module emphasizes the dependencies in the channels and computes the channel attention maps. Finally, the outputs of the two attention modules are fused and passed to convolutional blocks to enhance the feature representation, leading to better segmentation masks.}
    \label{fig:Architecture}
    \vspace{-0.5cm}
\end{figure*}

\noindent\textbf{Motivation}: To motivate our approach, we first distinguish two desirable properties to be considered when designing a hybrid framework that is efficient yet accurate.

\noindent\textbf{\textit{Efficient Global Attention}}: As discussed earlier, most existing hybrid methods employ self-attention operation having quadratic complexity in terms of the number of tokens. This is computationally expensive in the case of volumetric medical segmentation and becomes more problematic when interleaving window attention and convolution components in hybrid designs. Different from these approaches, we argue that computing self-attention across feature channels instead of volume dimension is expected to reduce the complexity from quadratic to linear with respect to the volumetric dimension. Further, the spatial attention information can be efficiently learned by projecting the spatial matrices of the keys and values into a lower-dimension space.

\noindent\textbf{\textit{Enriched Spatial-channel Feature Representation}}: Most existing hybrid volumetric medical image segmentation approaches typically capture the spatial features through attention computation and ignore the channel information in the form of encoding the inter-dependencies between different channel feature maps. Effectively combining the interactions in the spatial dimensions and the inter-dependencies between the channel features is expected to provide enriched contextual spatial-channel feature representations, leading to improved mask predictions.

\subsection{Overall Architecture}
\label{sec:Overall Architecture}
\noindent Fig.~\ref{fig:Architecture} presents our UNETR++ architecture, comprising a hierarchical encoder-decoder structure. We base our UNETR++ framework on the recently introduced UNETR~\cite{UNETR}  with skip connections between the encoders and decoders, followed by convolutional blocks (ConvBlocks) to generate the prediction masks. Instead of using a fixed feature resolution throughout the encoders, our UNETR++ employs a hierarchical design where the resolution of the features is gradually decreased by a factor of two in each stage.
Within our UNETR++ framework, the encoder has four stages, where the first stage consists of patch embedding to divide volumetric input into 3D patches, followed by our novel efficient paired-attention (EPA) block. In the patch embedding, we divide each 3D input (volume) $\bm{x} \in \mathbb{R}^{H \times W \times D}$ into non-overlapping patches $\bm{x}_u \in \mathbb{R}^{N \times (P_1, P_2, P_3)}$, where ($P_1, P_2, P_3$) is the resolution of each patch and $N = (\frac{H}{P_1} \times \frac{W}{P_2} \times \frac{D}{P_3})$ denotes the length of the sequence. Then, the patches are projected into $C$ channel dimensions, producing feature maps of size $\frac{H}{P_1} \times \frac{W}{P_2} \times \frac{D}{P_3} \times C$. We use the same patch resolution (4, 4, 2), as in ~\cite{nnFormer}. For each of the remaining encoder stages, we employ downsampling layers using non-overlapping convolution to decrease the resolution by a factor of two, followed by the EPA block.

Within our proposed UNETR++ framework, each EPA block comprises two attention modules to efficiently learn enriched spatial-channel feature representations by encoding the information in both spatial and channel dimensions with shared keys-queries scheme. The encoder stages are connected with the decoder stages via skip connections to merge the outputs at different resolutions. This enables the recovery of the spatial information lost during the downsampling operations, leading to predicting a more precise output. Similar to the encoder, the decoder also comprises four stages, where each decoder stage consists of an upsampling layer using deconvolution to increase the resolution of the feature maps by a factor of two, followed by the EPA block (except the last decoder). The number of channels is decreased by a factor of two between each two decoder stages. Consequently, the outputs of the last decoder are fused with convolutional features maps to recover the spatial information and enhance the feature representation. The resulting output is then fed into 3$\times$3$\times$3 and 1$\times$1$\times$1 convolutional blocks to generate voxel-wise final mask predictions. Next, we present in detail our EPA block.

\subsection{Efficient Paired-Attention Block}
\label{sec:EPA}
\noindent The proposed EPA block performs efficient global attention and effectively captures enriched spatial-channel feature representations. The EPA block comprises spatial attention and channel attention modules. The spatial attention module reduces the complexity of the self-attention from quadratic to linear. On the other hand, the channel attention module effectively learns the inter-dependencies between the channel feature maps. The EPA block is based on a shared keys-queries scheme between the two attention modules to be mutually informed in order to generate better and more efficient feature representation. 
This is likely due to learning complementary features by sharing the keys and queries but using different value layers. 

As illustrated in Fig.~\ref{fig:Architecture} (right), the input feature maps $\bm{x}$ are fed into the channel and spatial attention modules of the EPA block.
The weights of $\bm{Q}$ and $\bm{K}$ linear layers are shared across the two attention modules and different $\bm{V}$ layer is used for each attention module. The two attention modules are computed as:
\begin{equation}
    \hat{\bm{X}_s} = \txt{SA}(\bm{Q}_{shared},~{\bm{K}_{shared}},~{\bm{V}_{spatial}}),
    \label{eq:spatial_attention}
\end{equation}
\begin{equation}
    \hat{\bm{X}_c} = \txt{CA}(\bm{Q}_{shared},~{\bm{K}_{shared}},~{\bm{V}_{channel}}),
    \label{eq:channel_attention}
\end{equation}
where, $\hat{\bm{X}_s}$ and $\hat{\bm{X}_c}$ denotes the spatial and channels attention maps, respectively. $\txt{SA}$ is the spatial attention module, and $\txt{CA}$ is the channel attention module. $\bm{Q}_{shared}$, $\bm{K}_{shared}$, $\bm{V}_{spatial}$, and $\bm{V}_{channel}$ are the matrices for shared queries, shared keys, spatial value layer, and channel value layer, respectively.\\ 
\noindent
\textbf{Spatial Attention}: We strive in this module to learn the spatial information efficiently by reducing the complexity from $O(n^2)$ to $O(np)$, where $n$ is the number of tokens, and $p$ is the dimension of the projected vector, where $p<<n$. Given a normalized tensor $\bm{X}$ of shape $HWD$${\times}$$C$, we compute ${\bm{Q}_{shared}}$, ${\bm{K}_{shared}}$, and ${\bm{V}_{spatial}}$ projections using three linear layers, yielding ${\bm{Q}_{shared}}{=}{\bm{W}}^Q{\bm{X}}$, ${\bm{K}_{shared}}{=}{\bm{W}}^K{\bm{X}}$, and ${\bm{V}_{spatial}}{=}{\bm{W}}^V{\bm{X}}$, with dimensions $HWD$${\times}$$C$, where ${\bm{W}}^Q$,${\bm{W}}^K$, and ${\bm{W}}^V$ are the projection weights for ${\bm{Q}_{shared}}$, ${\bm{K}_{shared}}$, and ${\bm{V}_{spatial}}$, respectively. Then, we perform three steps. First, the $\bm{K}_{shared}$ and $\bm{V}_{spatial}$ layers are projected from $HWD \times C$ into lower-dimensional matrices of shape $p \times C$. Second, the spatial attention maps are computed by multiplying the $\bm{Q}_{shared}$ layer by the transpose of the projected $\bm{K}_{shared}$, followed by softmax to measure the similarity between each feature and the rest of the spatial features. Third, these similarities are multiplied by the projected $\bm{V}_{spatial}$ layer to produce the final spatial attention maps of shape $HWD\times C$. The spatial attention is defined as follows:
\begin{equation}
    \hat{\bm{X}_s} = \txt{Softmax}\Big(\frac{\bm{Q}_{shared} \bm{K}_{proj}^\top}{\sqrt{d}}\Big)\cdot\bm{\tilde{V}}_{spatial},
    \label{eq:SA}
\end{equation}
where, $\bm{Q}_{shared}$, $\bm{K}_{proj}$, $\bm{\tilde{V}}_{spatial}$ denote shared queries, projected shared keys, and projected spatial value layer, respectively, and $d$ is the size of each vector.

\noindent
\textbf{Channel Attention}:
This module captures the inter-dependencies between feature channels by applying the dot-product operation in the channel dimension between channel value layer and channel attention maps. Using the same $\bm{Q}_{shared}$ and $\bm{K}_{shared}$ of the spatial attention module, we compute value layer for the channels to learn the complementary features using linear layer, yielding ${\bm{V}_{channel}}={\bm{W}}^V{\bm{X}}$, with dimensions $HWD$${\times}$$C$, where ${\bm{W}}^V$ is the projection weight for ${\bm{V}_{channel}}$. The channel attention is defined as follows:
 \begin{equation}
    \hat{\bm{X}_c}=\bm{V}_{channel}\cdot\txt{Softmax}\Big(\frac{\bm{Q}_{shared}^\top \bm{K}_{shared}}{\sqrt{d}}\Big),
    \label{eq:CA}
\end{equation}
where, $\bm{V}_{channel}$, $\bm{Q}_{shared}$, $\bm{K}_{shared}$ denote channel value layer, shared queries, and shared keys, respectively, and $d$ is the size of each vector.

Finally, we perform sum fusion and transform the outputs from the two attention modules by convolution blocks to obtain enriched feature representations. The final output $\hat{X}$ of the EPA block is obtained as:
\begin{equation}
    \hat{\bm{X}} = \txt{Conv}_{1}(\txt{Conv}_{3}(\hat{\bm{X}_s} + \hat{\bm{X}_c})),
    \label{eq:Paired_attention}
\end{equation}
where, $\hat{\bm{X}_s}$ and $\hat{\bm{X}_c}$ denotes the spatial and channels attention maps, and $\txt{Conv}_{1}$ and $\txt{Conv}_{3}$ are $1\times1\times1$ and $3\times3\times3$ convolution blocks, respectively.

\subsection{Loss Function}
Our loss function is based on a summation of the commonly used soft dice loss~\cite{milletari2016v} and cross-entropy loss to simultaneously leverage the benefits of both complementary loss functions. It is defined as: 
\begin{equation}
\begin{split}
    \mathcal{L}(Y,P) = 1-\sum_{i=1}^{I}\ & \left(\frac{2*\sum_{v=1}^{V} Y_{v,i} \cdot {P_{v,i}}}{\sum_{v=1}^{V}Y_{v,i}^2 + \sum_{v=1}^{V}{P_{v,i}^2}} \right.\\ 
    & +  \left. \sum_{v=1}^{V}Y_{v,i} \log {P_{v,i}} \right),
    \label{eq:Loss}
\end{split}
\end{equation}
where, $I$ denotes the number of classes;  $V$ denotes the number of voxels; $Y_{v,i}$ and ${P_{v,i}}$ denote the ground truths and output probabilities at voxel $v$ for class $i$, respectively.
\section{Experiments}

\begin{table*}[!ht]
    \begin{center}
        \resizebox{\textwidth}{!}{
        \begin{tabular}{l|c c| c c c c c c c c|cc}
        \toprule
            \multirow{2}{*}{Methods} & \multirow{2}{*}{Params} & \multirow{2}{*}{FLOPs} & \multirow{2}{*}{Spl} &  \multirow{2}{*}{RKid} &  \multirow{2}{*}{ LKid} & \multirow{2}{*}{Gal}  & \multirow{2}{*}{Liv}  & \multirow{2}{*}{Sto} & \multirow{2}{*}{Aor} &  \multirow{2}{*}{Pan} &  \multicolumn{2}{c}{Average} 
            
            \\ \cmidrule{12-13}
             & & & & & & & & & & & HD95 $\downarrow$ & DSC $\uparrow$ \\
                \midrule
                \midrule
        U-Net~\cite{UNet} & - & - & 86.67 & 68.60 & 77.77& 69.72 & 93.43 & 75.58  &  89.07  & 53.98 & - & 76.85 \\
        TransUNet~\cite{TransUNet} & 96.07 & 88.91 & 85.08 & 77.02 & 81.87 & 63.16 & 94.08 & 75.62 &  87.23  & 55.86 & 31.69 & 77.49 \\
        
        Swin-UNet~\cite{cao2021swin} & - & - & 90.66 &  79.61 & 83.28 & 66.53 & 94.29 & 76.60 & 85.47 &   56.58  & 21.55 & 79.13   \\
        UNETR~\cite{UNETR} & 92.49 & 75.76 & 85.00 & 84.52 & 85.60 & 56.30 & 94.57 & 70.46 & 89.80 & 60.47 &  18.59 & 78.35  \\
        MISSFormer~\cite{MISSFormer} & - & - & 91.92 &  82.00 & 85.21 &  68.65 & 94.41 & 80.81 & 86.99 & 65.67 & 18.20 & 81.96 \\
        
        Swin-UNETR~\cite{SWIN_UNETR} & 62.83 & 384.2 & 95.37 & 86.26 & 86.99 & 66.54 & 95.72 & 77.01 & 91.12 &  68.80 &  10.55 & 83.48 \\
        
        nnFormer~\cite{nnFormer} & 150.5 & 213.4 & 90.51 & 86.25 & 86.57 & 70.17 &\textbf{96.84} & \textbf{86.83}  & 92.04 &    \textbf{83.35} &  10.63 & 86.57\\
        \midrule
        \rowcolor{orange!6} UNETR++ & \textbf{42.96} & \textbf{47.98} & \textbf{95.77} & \textbf{87.18} & \textbf{87.54} & \textbf{71.25} & 96.42 & 86.01 & \textbf{92.52} & 81.10  &  \textbf{7.53} & \textbf{87.22} \\
        \bottomrule
        \end{tabular}
        }\vspace{-0.5em}
        \caption{State-of-the-art comparison on the abdominal multi-organ Synapse dataset. We report both the segmentation performance (DSC, HD95) and model complexity (parameters and FLOPs).
         Our proposed UNETR++ achieves favorable segmentation performance against existing methods, while being considerably reducing the model complexity. Best results are in bold. Abbreviations stand for: Spl: \textit{spleen}, RKid: \textit{right kidney}, LKid: \textit{left kidney}, Gal: \textit{gallbladder}, Liv: \textit{liver}, Sto: \textit{stomach}, Aor: \textit{aorta}, Pan: \textit{pancreas}. Best results are in bold.}
        \label{table:synapse}
        \end{center}
\vspace{-0.8cm}
\end{table*}


\subsection{Experimental Setup}
\noindent
We carry out experiments on five datasets: Synapse for Multi-organ CT Segmentation~\cite{BTCV}, BTCV for Multi-organ CT Segmentation~\cite{BTCV}, ACDC for Automated Cardiac Diagnosis~\cite{ACDC},  Brain Tumor Segmentation (BraTS)~\cite{BRATS} and the Medical Segmentation Decathlon-Lung~\cite{MSD}.\\
\textbf{Datasets:}
The \textit{Synapse}~\cite{BTCV} dataset consists of abdominal CT scans of 30 subjects with 8 organs. Consistent with previous approaches,
we follow the splits used in~\cite{TransUNet} and train our model on 18 samples and evaluate on the remaining 12 cases. We report the model performance using Dice Similarity Coefficient (DSC) and 95\% Hausdorff Distance (HD95) on 8 abdominal organs: \textit{spleen}, \textit{right kidney}, \textit{left kidney}, \textit{gallbladder}, \textit{liver}, \textit{stomach}, \textit{aorta} and \textit{pancreas}. The
\textit{BTCV}~\cite{BTCV} dataset contains 30 subjects for training and 20 subjects for testing with abdominal CT scans. It consists of 13 organs, including 8 organs of Synapse, along with \textit{esophagus}, \textit{inferior vena cava}, \textit{portal and splenic veins}, \textit{right and left adrenal gland}. We report the DSC on all 13 abdominal organs. The \textit{ACDC}~\cite{ACDC} dataset comprises cardiac MRI images of 100 patients, with segmentation annotations of \textit{right ventricle} (RV), \textit{left ventricle} (LV) and \textit{myocardium} (MYO).

Consistent with~\cite{nnFormer}, we split the data into 70, 10 and 20 train, validation and test samples. We report the DSC on the three classes. The \textit{BraTS}~\cite{BRATS} comprises of 484 MRI images, where each image consists of four channels, FLAIR, T1w, T1gd and T2w. We split the dataset into 80:5:15 ratio for training, validation and testing and report on the test set. The target categories are whole tumor, enhancing tumor and tumor core. The lung~\cite{MSD} dataset comprises 63 CT volumes for a two-class problem with the goal to segment lung cancer from the background. We split the data into 80:20 ratio for training and validation.\\
\noindent
\textbf{Evaluation Metrics:} We measure the performance of the models based on two metrics: Dice Similarity Coefficient (DSC) and 95\% Hausdorff Distance (HD95). DSC measures the overlap between the volumetric segmentation predictions and the voxels of the ground truths, it is defined as follows:
\begin{equation}
    {DSC}(Y,P) =  2 * \frac{ |Y \cap P|} {|Y| \cup |P|}=  2 * \frac{Y \cdot P} {Y^2 + P^2}
    \label{eq:DSC}
\end{equation}
where, $Y$ and $P$ denote the ground truths and output probabilities for all voxels, respectively.

HD95 is commonly used as boundary-based metric to measure the $95^{th}$ percentile of the distances between boundaries of the volumetric segmentation predictions and the voxels of the ground truths. It is defined as follows:
\begin{equation}
    {HD_{95}}(Y,P) = \max \{ \bm{d}_{YP}, \bm{d}_{PY} \}
    \label{eq:HD95}
\end{equation}
where, $\bm{d}_{YP}$ is the maximum $95^{th}$ percentile distance between predicted voxels and the ground truth, and $\bm{d}_{PY}$ is the maximum $95^{th}$ percentile distance between the ground truth and the predicted voxels.

\noindent
\textbf{Implementation Details:}
We implement our approach in Pytorch v1.10.1 and using the MONAI libraries~\cite{monai}. For a fair comparison with both the baseline UNETR and nnFormer, we use the same input size, pre-processing strategy and no additional training data. The models are trained using a single A100 40GB GPU with input 3D patches of size $128\times128\times64$ for 1k epochs with learning rate of 0.01 and weight decay of 3e$^{-5}$. In addition, we report results with $96\times96\times96$ input size and patch resolution of (4, 4, 4) for BTCV where the models are trained for 5k epochs with learning rate of 1e$^{-4}$. Specifically, the input volume is divided into non-overlapping patches during training which are used to learn segmentation maps through back-propagation. During training, we apply the same data augmentations for UNETR, nnFormer and our UNETR++. More details are provided in the suppl. material.
\begin{table}[!ht]
\centering
\resizebox{\columnwidth}{!}{
\begin{tabular}{l *{3}{c}}
  \toprule
 Model & Params (M) & FLOPs (G) & DSC (\%) \\
 \midrule
 \midrule
   UNETR (Baseline) & 92.49 & 75.76 & 78.35 \\
   + EPA in Encoder    & 28.94 & 39.36 & 85.17 \\
        \rowcolor{orange!6}+ EPA in Decoder (UNETR++)    & 42.96 & 47.98 & 87.22 \\
  \bottomrule                             
\end{tabular}
}
\caption{Baseline comparison on Synapse.
We show the results in terms of segmentation performance (DSC) and model complexity (parameters and FLOPs). For a fair comparison, all results are obtained using the same input size and pre-processing. Integrating the EPA block in the encoders of our hierarchical design improves the segmentation performance to 85.17\%. The results are further improved to 87.22\% by also introducing the EPA block in decoders. Our UNETR++ with the novel EPA block both in the encoders and decoders achieves an absolute gain of 8.87\% in DSC, while also significantly reducing the model complexity.}
\label{table1_baseline}
\vspace{-0.3cm}
\end{table}
\begin{table*}[h]
\resizebox{\textwidth}{!}{
\begin{tabular}{l|rrrrrrrrrrrrr|r}
\midrule
Methods   & \multicolumn{1}{l}{Spl}          & \multicolumn{1}{l}{RKid}   & \multicolumn{1}{l}{LKid}   & \multicolumn{1}{l}{Gal}            & \multicolumn{1}{l}{Eso}    & \multicolumn{1}{l}{Liv}  & \multicolumn{1}{l}{Sto}   & \multicolumn{1}{l}{Aor}  & \multicolumn{1}{l}{IVC}    & \multicolumn{1}{l}{PSV}    & \multicolumn{1}{l}{Pan}   & \multicolumn{1}{l}{RAG}    & \multicolumn{1}{l|}{LAG}    & \multicolumn{1}{l}{Avg}    \\ \midrule \midrule
nnUNet~\cite{nnUNet}
& \textbf{95.95} & 88.35 & 93.02 & 70.13 & 76.72 & \textbf{96.51} & \textbf{86.79} & 88.93 & 82.89 & \textbf{78.51} & \textbf{79.60} & \textbf{73.26} & \textbf{68.35} & 83.16 \\
TransBTS~\cite{TransBTS}
& 94.55  & 89.20 & 90.97  & 68.38  & 75.61  & 96.44  & 83.52  & 88.55 & 82.48 & 74.21 & 76.02 & 67.23 & 67.03 & 81.31  \\
UNETR~\cite{UNETR}
& 90.48  & 82.51 & 86.05  & 58.23  & 71.21  & 94.64  & 72.06  & 86.57  & 76.51  & 70.37  & 66.06 & 66.25 & 63.04 & 76.00 \\
Swin-UNETR~\cite{SWIN_UNETR}
& 94.59  & 88.97 & 92.39  & 65.37  & 75.43  & 95.61  & 75.57  & 88.28  & 81.61  & 76.30  & 74.52 & 68.23 & 66.02 & 80.44                     \\ 
nnFormer~\cite{nnFormer}
& 94.58  & 88.62 & \textbf{93.68} & 65.29  & 76.22  & 96.17  & 83.59  & 89.09 & 80.80 & 75.97 & 77.87 & 70.20 & 66.05 & 81.62 \\
\midrule
\rowcolor{orange!6} UNETR++  & 94.94 & \textbf{91.90} & 	
93.62 & \textbf{70.75} & \textbf{77.18} & 95.95 & 85.15 & \textbf{89.28}  & \textbf{83.14} & 76.91 & 77.42 & 72.56 & 68.17 & \textbf{83.28} \\
\midrule
\end{tabular}
}
\caption{State-of-the-art comparison on the BTCV test set for multi-organ segmentation. All results are obtained using a single model accuracy and without any ensemble, pre-training or additional custom data. Our UNETR++ achieves favorable segmentation performance against existing 3D image segmentation methods. Abbreviations are as follows: Spl: \textit{spleen}, RKid: \textit{right kidney}, LKid: \textit{left kidney}, Gal: \textit{gallbladder}, Eso: \textit{esophagus}, Liv: \textit{liver}, Sto: \textit{stomach}, Aor: \textit{aorta}, IVC: \textit{the inferior vena cava}, PSV: \textit{portal and splenic veins}, Pan: \textit{pancreas}, RAG: \textit{right adrenal gland}, LAG: \textit{left adrenal gland}. Results are obtained from BTCV leaderboard.}
\label{tab:btcv}
\end{table*}

\subsection{Baseline Comparison}
\begin{figure}[t]

\centering
{\includegraphics[width=0.475\textwidth]{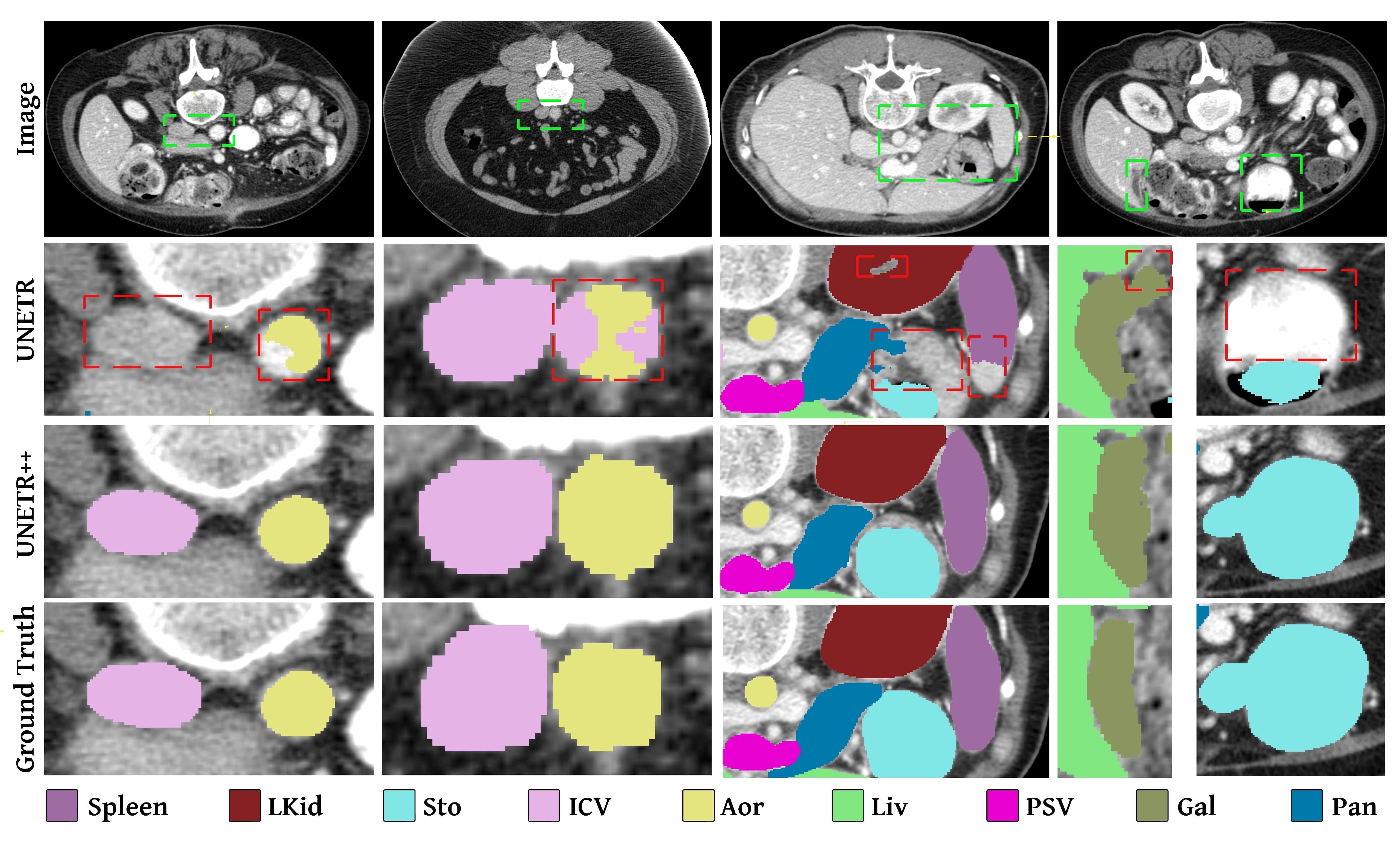}}
\caption{Qualitative comparison between UNETR++ and baseline UNETR on Synapse. For better visualization, we enlarged different areas (marked in green dashed box) in the images. The inaccurate segmentations are marked by red dashed boxes. Compared to the baseline, UNETR++ achieves superior segmentation performance.
Best viewed in zoom.}
\label{fig:Baseline_comparison}
\vspace{-5mm}
\end{figure}

\noindent Tab.~\ref{table1_baseline} shows the impact of integrating the proposed contributions within the baseline UNETR~\cite{UNETR} on Synapse. In addition to the Dice Similarity Coefficient (DSC), we report the model complexity in terms of parameters and FLOPs. In all cases, we report performance in terms of single model accuracy. As discussed earlier, UNETR++ is a hierarchical architecture that downsamples the feature maps of the encoder by a factor of two after each stage. Hence, the model comprises four encoder stages and four decoder stages. This hierarchical design of our UNETR++ enables a significant reduction in model complexity by reducing the parameters from 92.49M to 16.60M and FLOPs from 75.76G to 30.75G while maintaining a comparable DSC of 78.29\%, compared to the baseline. Introducing the EPA block within our UNETR++ encoders leads to a significant improvement in performance with an absolute gain of 6.82\% in DSC over the baseline. The performance is further improved by integrating the EPA block in the decoder. Our final UNETR++ having a hierarchical design with the novel EPA block both in encoders and decoders leads to a significant improvement of 8.87\% in DSC, while considerably reducing the model complexity by 54\% in parameters and 37\% in FLOPs, compared to the baseline. 
We further conduct an experiment to evaluate our spatial and channel attention within the proposed EPA block. Employing spatial and channel attention improve the performance significantly with DSC of 86.42\% and 86.39\%, respectively over the baseline. 
Combining both spatial and channel attention within our EPA block leads to a further improvement with DSC of 87.22\%.
Fig.~\ref{fig:Baseline_comparison} shows a qualitative comparison between the baseline and our UNETR++ on the Synapse dataset. 
We enlarge different organs (marked as green dashed boxes in the first row) from several cases. In column 1, the baseline struggles to segment the \textit{inferior vena cava} and \textit{aorta}.
In column 2, it confuses the same two organs when they are adjacent to each other. In the last two columns, the baseline under-segments \textit{left kidney}, \textit{spleen}, and \textit{stomach}, whereas it over-segments the \textit{gallblader}. In contrast, UNETR++ achieves improved performance by accurately segmenting all organs. 

\begin{figure*}[!ht]
\centering
{\includegraphics[width=1.0\textwidth]{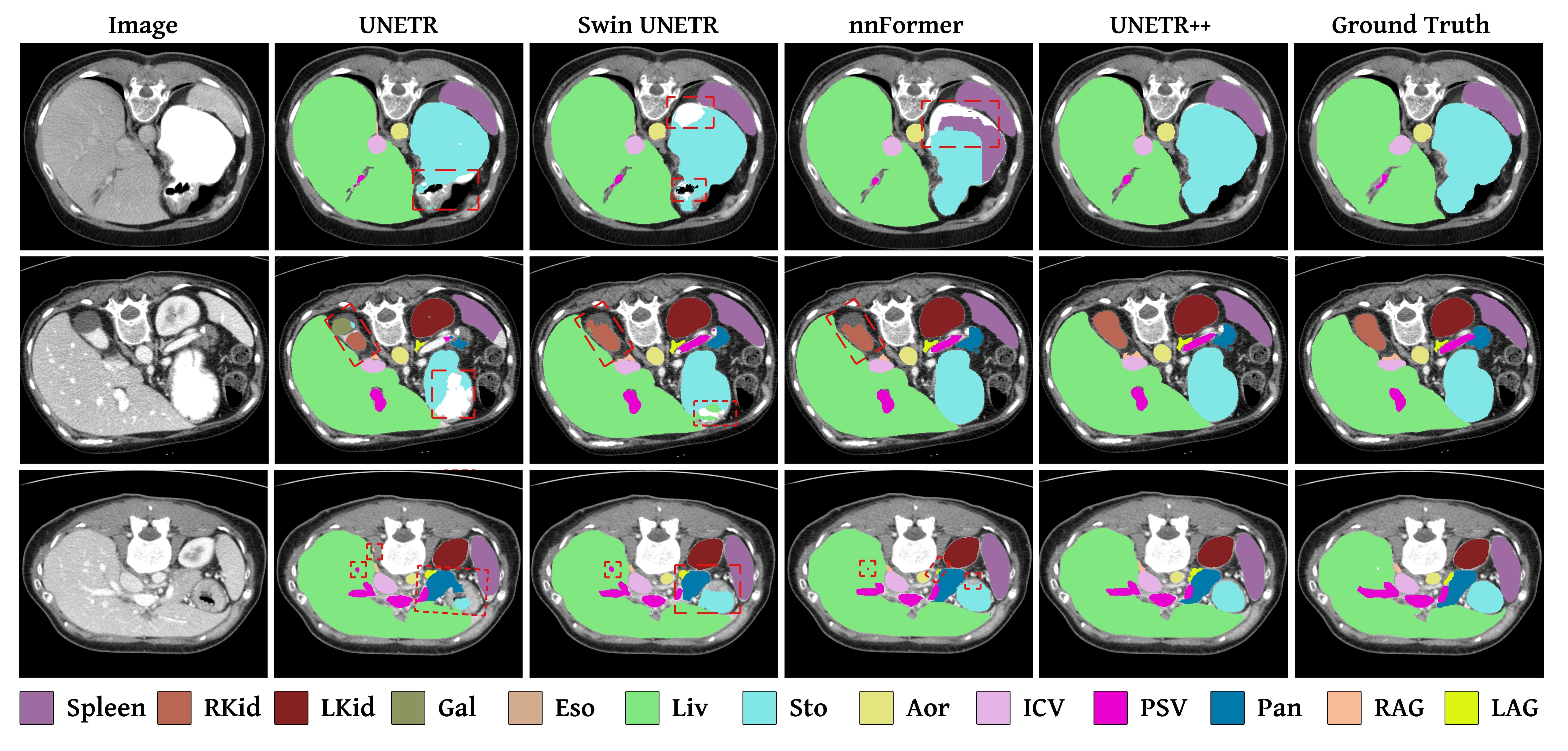}}
\vspace{-2em}
\caption{Qualitative comparison on multi-organ segmentation task. Here, we compare our UNETR++ with existing methods: UNETR, Swin UNETR, and nnFormer. Existing methods struggle to correctly segment different organs (marked in red dashed box). Our UNETR++ achieves promising segmentation performance by accurately segmenting the organs. Best viewed in zoom.}
\label{Fig:Qualitative_results}
\vspace{-0.2cm}
\end{figure*}

\subsection{State-of-the-Art Comparison}
\noindent
\textbf{Synapse Dataset:} Tab.~\ref{table:synapse} shows the results on the multi-organ Synapse dataset. We report the segmentation performance using DSC and HD95 metrics on the abdominal organs. In addition, we report the model complexity in terms of parameters and FLOPs for each method. The segmentation performance is reported with a single model accuracy and without utilizing any pre-training, model ensemble or additional data. The pure CNN-based U-Net~\cite{UNet} approach achieves a DSC of 76.85\%. Among existing hybrid transformers-CNN based methods, UNETR~\cite{UNETR} and Swin-UNETR~\cite{SWIN_UNETR} achieve DSC of 78.35\% and 83.48\%, respectively. On this dataset, nnFormer~\cite{nnFormer} obtains superior performance compared to other existing works. Our UNETR++ outperforms nnFormer by achieving a DSC of 87.22\%. Further, UNETR++ obtains an absolute reduction in error of 3.1\% over nnFormer in terms of HD95 metric. Notably, UNETR++ achieves this improvement in segmentation performance by significantly reducing the model complexity by over 71\% in terms of parameters and FLOPs.

Fig.~\ref{Fig:Qualitative_results} shows a qualitative comparison of UNETR++ with existing approaches on abdominal multi-organ segmentation. Here, the inaccurate segmentations are marked with red dashed boxes. In the first row, we observe 
that existing approaches struggle to accurately segment the \textit{stomach} by either under-segment it in the case of UNETR and Swin UNETR or confusing it with \textit{spleen} in the case of nnFormer. In comparison, our UNETR++ accurately segments the \textit{stomach}. Further, existing methods fail to fully segment the \textit{right kidney} in the second row. In contrast, our UNETR++ accurately segments the whole \textit{right kidney}, likely due to learning the contextual information with the enriched spatial-channel representation. Moreover, UNETR++ smoothly delineates boundaries between \textit{spleen}, \textit{stomach} and \textit{liver}. In the third row, UNETR confuses \textit{stomach} with \textit{pancreas}. On the other hand, Swin UNETR and nnFormer under-segment \textit{stomach} and \textit{left adrenal gland}, respectively. UNETR++ accurately segments all organs with better delineation of the boundaries in these examples. 

\noindent
\textbf{BTCV Dataset:} Tab.~\ref{tab:btcv} presents the comparison on BTCV test set. Here, all results are based on a single model accuracy without any ensemble, pre-training or additional data. We report results on all 13 organs along with corresponding mean performance over all organs. Among existing works, UNETR and SwinUNETR achieve a mean DSC of 76.0\% and 80.44\%. Among existing methods, nnUNet obtains a performance of 83.16\% mean DSC, but requires 358G FLOPs. In comparison, UNETR++ performs favorably against nnUNet by achieving a mean DSC of 83.28\%, while requiring significantly fewer FLOPs of 31G. 

\begin{table}[t]
\centering
        \begin{tabular}{l|ccc|c}
        \toprule
        Methods & RV & Myo & LV & Average  \\
                \midrule
                \midrule
        TransUNet~\cite{TransUNet} 
        & 88.86  & 84.54 & 95.73 & 89.71 \\
        Swin-UNet~\cite{cao2021swin} 
        & 88.55  & 85.62 & 95.83 & 90.00 \\
        UNETR~\cite{UNETR} 
        & 85.29 & 86.52 & 94.02 & 86.61 \\
        MISSFormer~\cite{MISSFormer} 
        & 86.36 & 85.75 & 91.59 & 87.90 \\
        nnFormer~\cite{nnFormer} 
        & 90.94 & 89.58 & 95.65 & 92.06 \\
        \midrule
        \rowcolor{orange!6} UNETR++ & \textbf{91.89} & \textbf{90.61} & \textbf{96.00} & \textbf{92.83} \\
        \bottomrule
        \end{tabular}
        \vspace{0.1em}
        \caption{State-of-the-art comparison on ACDC. We report the performance on \textit{right ventricle} (RV), \textit{left ventricle} (LV) and \textit{myocardium} (MYO) along with mean results using DSC metric.
        }
        \label{table:ACDC}
\vspace{-0.1cm}
\end{table}
\begin{table}[t]
\centering
\resizebox{\linewidth}{!}{%
        \begin{tabular}{lccccc|c}
\toprule
Model & {Params} & {FLOPs} & {Mem} & {GPU T.} & {CPU T.} & {DSC (\%)}   \\
\midrule
\midrule
UNETR~\cite{UNETR}  & 92.5 & 153.5 & 3.3 & 82.5 & 2145.0 & 81.2   \\
SwinUNETR~\cite{SWIN_UNETR} & 62.8 & 572.4 & 19.7 & 228.6 & 7612.3 & 81.5   \\
nnFormer~\cite{nnFormer} & 149.6 & 421.5 & 12.6 &  148.0 & 5247.5 & 82.3  \\
\midrule
\rowcolor{orange!6}\textbf{UNETR++} & \textbf{42.6} & \textbf{70.1} & \textbf{2.4} & \textbf{62.4} & \textbf{1497.7} & \textbf{82.8}   \\
\bottomrule
\end{tabular}}
        \vspace{0.1em}
        \caption{Comparison on BRaTs. UNETR++ achieves favorable segmentation results (DSC), while being \textit{efficient} (Params in millions and GFLOPs), operating at \textit{faster inference speed} (GPU T. and CPU T. in ms) and requires \textit{lesser GPU memory} (Mem in GB).}
        \label{table:brats}
\vspace{-0.5cm}
\end{table}
\noindent
\textbf{ACDC Dataset:} Tab.~\ref{table:ACDC}  shows the comparison on ACDC. Here, all results are reported with a single model accuracy and without using any pre-training, model ensemble or additional data. UNETR and nnFormer achieve mean DSC of 86.61\% and 92.06\%, respectively. UNETR++ achieves improved performance with a mean DSC of 92.83\%.\\
\noindent \textbf{BRaTs Dataset:} Tab.~\ref{table:brats} shows segmentation performance, model complexity, and inference time. For a fair comparison, we use same input size and pre-processing strategy. We compare speed on Quadro RTX 6000 24 GB GPU \& 32 Core Intel(R) Xeon(R) 4215 CPU. Here, inference time is avg. forward pass time using {1$\times$128$\times$128$\times$128} input size of BRaTs. Compared to recent transformer-based methods, our UNETR++ achieves favourable performance while operating at a \textit{faster inference speed} as well as requiring significantly \textit{lesser GPU memory}. 

\begin{wraptable}{r}{3.7cm}
\vspace{-0.2in}
\setlength{\tabcolsep}{2pt}
\resizebox{1.0\linewidth}{!}{
  \begin{tabular}{lc}
  \toprule
 Model & DSC (\%) \\
  \midrule
  \midrule
nnUNet~\cite{nnUNet} & 74.31\\
SwinUNETR~\cite{SWIN_UNETR}  & 75.55\\
nnFormer~\cite{nnFormer}   & 77.95\\
UNETR~\cite{UNETR}  & 73.29\\
\midrule
\rowcolor{orange!6}UNETR++ & \textbf{80.68}\\						
  \bottomrule
\end{tabular}
}

\label{tab:lungs}
\vspace{-0.2cm}
\end{wraptable}
\noindent \textbf{Lungs Dataset:}
We evaluate UNETR++ and other SOTA models on the lung cancer segmentation task. UNETR++ obtains better segmentation performance compared to existing methods by achieving a mean DSC of 80.68\%.

\section{Conclusion}
\noindent We propose a hierarchical approach, named UNETR++, for 3D medical segmentation. Our UNETR++ introduces an efficient paired attention (EPA) block to encode enriched inter-dependent spatial and channel features by using spatial and channel attention. Within the EPA block, we share the weights of query and key mapping functions to better communicate between spatial and channel branches, providing complementary benefits as well as reducing the parameters. 
Our UNETR++ achieves favorable segmentation results on five datasets while significantly reducing the model complexity with better speed, compared to existing methods.




{\small
\bibliographystyle{ieee_fullname}
\bibliography{egbib}

\begin{thebibliography}{10}\itemsep=-1pt

\bibitem{ACDC}
Olivier Bernard, Alain Lalande, Clement Zotti, Frederick Cervenansky, Xin Yang,
  Pheng-Ann Heng, Irem Cetin, Karim Lekadir, Oscar Camara, Miguel~Angel
  Gonzalez~Ballester, Gerard Sanroma, Sandy Napel, Steffen Petersen, Georgios
  Tziritas, Elias Grinias, Mahendra Khened, Varghese~Alex Kollerathu, Ganapathy
  Krishnamurthi, Marc-Michel Rohé, Xavier Pennec, Maxime Sermesant, Fabian
  Isensee, Paul Jäger, Klaus~H. Maier-Hein, Peter~M. Full, Ivo Wolf, Sandy
  Engelhardt, Christian~F. Baumgartner, Lisa~M. Koch, Jelmer~M. Wolterink,
  Ivana Išgum, Yeonggul Jang, Yoonmi Hong, Jay Patravali, Shubham Jain,
  Olivier Humbert, and Pierre-Marc Jodoin.
\newblock Deep learning techniques for automatic mri cardiac multi-structures
  segmentation and diagnosis: Is the problem solved?
\newblock {\em IEEE Transactions on Medical Imaging}, 37(11):2514--2525, 2018.

\bibitem{cai2020dense}
Sijing Cai, Yunxian Tian, Harvey Lui, Haishan Zeng, Yi Wu, and Guannan Chen.
\newblock Dense-unet: a novel multiphoton in vivo cellular image segmentation
  model based on a convolutional neural network.
\newblock {\em Quantitative Imaging in Medicine and Surgery}, 10(6):1275, 2020.

\bibitem{cao2021swin}
Hu Cao, Yueyue Wang, Joy Chen, Dongsheng Jiang, Xiaopeng Zhang, Qi Tian, and
  Manning Wang.
\newblock Swin-unet: Unet-like pure transformer for medical image segmentation.
\newblock In {\em European Conference on Computer Vision Workshops}, 2022.

\bibitem{DETR}
Nicolas Carion, Francisco Massa, Gabriel Synnaeve, Nicolas Usunier, Alexander
  Kirillov, and Sergey Zagoruyko.
\newblock End-to-end object detection with transformers.
\newblock In {\em European Conference on Computer Vision}, 2020.

\bibitem{TransUNet}
Jieneng Chen, Yongyi Lu, Qihang Yu, Xiangde Luo, Ehsan Adeli, Yan Wang, Le Lu,
  Alan~L Yuille, and Yuyin Zhou.
\newblock Transunet: Transformers make strong encoders for medical image
  segmentation.
\newblock {\em arXiv preprint arXiv:2102.04306}, 2021.

\bibitem{chen2018encoder}
Liang-Chieh Chen, Yukun Zhu, George Papandreou, Florian Schroff, and Hartwig
  Adam.
\newblock Encoder-decoder with atrous separable convolution for semantic image
  segmentation.
\newblock In {\em European Conference on Computer Vision}, 2018.

\bibitem{SparseAttention}
Rewon Child, Scott Gray, Alec Radford, and Ilya Sutskever.
\newblock Generating long sequences with sparse transformers.
\newblock {\em arXiv preprint arXiv:1904.10509}, 2019.

\bibitem{cciccek20163d}
{\"O}zg{\"u}n {\c{C}}i{\c{c}}ek, Ahmed Abdulkadir, Soeren~S Lienkamp, Thomas
  Brox, and Olaf Ronneberger.
\newblock 3d u-net: learning dense volumetric segmentation from sparse
  annotation.
\newblock In {\em International Conference on Medical Image Computing and
  Computer-Assisted Intervention}, 2016.

\bibitem{ViTs}
Alexey Dosovitskiy, Lucas Beyer, Alexander Kolesnikov, Dirk Weissenborn,
  Xiaohua Zhai, Thomas Unterthiner, Mostafa Dehghani, Matthias Minderer, Georg
  Heigold, Sylvain Gelly, et~al.
\newblock An image is worth 16x16 words: Transformers for image recognition at
  scale.
\newblock {\em arXiv preprint arXiv:2010.11929}, 2020.

\bibitem{dou20163d}
Qi Dou, Hao Chen, Yueming Jin, Lequan Yu, Jing Qin, and Pheng-Ann Heng.
\newblock 3d deeply supervised network for automatic liver segmentation from ct
  volumes.
\newblock In {\em International Conference on Medical Image Computing and
  Computer-Assisted Intervention}, 2016.

\bibitem{gibson2018automatic}
Eli Gibson, Francesco Giganti, Yipeng Hu, Ester Bonmati, Steve Bandula,
  Kurinchi Gurusamy, Brian Davidson, Stephen~P Pereira, Matthew~J Clarkson, and
  Dean~C Barratt.
\newblock Automatic multi-organ segmentation on abdominal ct with dense
  v-networks.
\newblock {\em IEEE transactions on medical imaging}, 37(8):1822--1834, 2018.

\bibitem{SWIN_UNETR}
Ali Hatamizadeh, Vishwesh Nath, Yucheng Tang, Dong Yang, Holger~R Roth, and
  Daguang Xu.
\newblock Swin unetr: Swin transformers for semantic segmentation of brain
  tumors in mri images.
\newblock In {\em International MICCAI Brainlesion Workshop}, 2022.

\bibitem{UNETR}
Ali Hatamizadeh, Yucheng Tang, Vishwesh Nath, Dong Yang, Andriy Myronenko,
  Bennett Landman, Holger~R Roth, and Daguang Xu.
\newblock Unetr: Transformers for 3d medical image segmentation.
\newblock In {\em Proceedings of the IEEE/CVF Winter Conference on Applications
  of Computer Vision}, 2022.

\bibitem{huang2020unet}
Huimin Huang, Lanfen Lin, Ruofeng Tong, Hongjie Hu, Qiaowei Zhang, Yutaro
  Iwamoto, Xianhua Han, Yen-Wei Chen, and Jian Wu.
\newblock Unet 3+: A full-scale connected unet for medical image segmentation.
\newblock In {\em IEEE International Conference on Acoustics, Speech and Signal
  Processing}, 2020.

\bibitem{MISSFormer}
Xiaohong Huang, Zhifang Deng, Dandan Li, and Xueguang Yuan.
\newblock Missformer: An effective medical image segmentation transformer.
\newblock {\em arXiv preprint arXiv:2109.07162}, 2021.

\bibitem{nnUNet}
Fabian Isensee, Paul~F Jaeger, Simon~AA Kohl, Jens Petersen, and Klaus~H
  Maier-Hein.
\newblock nnu-net: a self-configuring method for deep learning-based biomedical
  image segmentation.
\newblock {\em Nature methods}, 18(2):203--211, 2021.

\bibitem{karimi2021convolution}
Davood Karimi, Serge~Didenko Vasylechko, and Ali Gholipour.
\newblock Convolution-free medical image segmentation using transformers.
\newblock In {\em International Conference on Medical Image Computing and
  Computer-Assisted Intervention}, 2021.

\bibitem{Reformer}
Nikita Kitaev, Lukasz Kaiser, and Anselm Levskaya.
\newblock Reformer: The efficient transformer.
\newblock In {\em International Conference on Learning Representations}, 2020.

\bibitem{BTCV}
Bennett Landman, Zhoubing Xu, J Igelsias, Martin Styner, T Langerak, and Arno
  Klein.
\newblock Miccai multi-atlas labeling beyond the cranial vault--workshop and
  challenge.
\newblock In {\em MICCAI Multi-Atlas Labeling Beyond Cranial Vault—Workshop
  Challenge}, 2015.

\bibitem{li2020pgd}
Ziqiang Li, Hong Pan, Yaping Zhu, and A~Kai Qin.
\newblock Pgd-unet: A position-guided deformable network for simultaneous
  segmentation of organs and tumors.
\newblock In {\em International Joint Conference on Neural Networks}, 2020.

\bibitem{lin2022ds}
Ailiang Lin, Bingzhi Chen, Jiayu Xu, Zheng Zhang, Guangming Lu, and David
  Zhang.
\newblock Ds-transunet: Dual swin transformer u-net for medical image
  segmentation.
\newblock {\em IEEE Transactions on Instrumentation and Measurement}, 2022.

\bibitem{Swin}
Ze Liu, Yutong Lin, Yue Cao, Han Hu, Yixuan Wei, Zheng Zhang, Stephen Lin, and
  Baining Guo.
\newblock Swin transformer: Hierarchical vision transformer using shifted
  windows.
\newblock In {\em Proceedings of the IEEE/CVF International Conference on
  Computer Vision}, 2021.

\bibitem{Maaz2022EdgeNeXt}
Muhammad Maaz, Abdelrahman Shaker, Hisham Cholakkal, Salman Khan, Syed~Waqas
  Zamir, Rao~Muhammad Anwer, and Fahad~Shahbaz Khan.
\newblock Edgenext: Efficiently amalgamated cnn-transformer architecture for
  mobile vision applications.
\newblock In {\em International Workshop on Computational Aspects of Deep
  Learning at 17th European Conference on Computer Vision (CADL2022)}.
  Springer, 2022.

\bibitem{BRATS}
Bjoern~H. Menze, Andras Jakab, Stefan Bauer, Jayashree Kalpathy-Cramer, Keyvan
  Farahani, Justin Kirby, Yuliya Burren, Nicole Porz, Johannes Slotboom, Roland
  Wiest, Levente Lanczi, Elizabeth Gerstner, Marc-André Weber, Tal Arbel,
  Brian~B. Avants, Nicholas Ayache, Patricia Buendia, D.~Louis Collins, Nicolas
  Cordier, Jason~J. Corso, Antonio Criminisi, Tilak Das, Hervé Delingette,
  Çağatay Demiralp, Christopher~R. Durst, Michel Dojat, Senan Doyle, Joana
  Festa, Florence Forbes, Ezequiel Geremia, Ben Glocker, Polina Golland,
  Xiaotao Guo, Andac Hamamci, Khan~M. Iftekharuddin, Raj Jena, Nigel~M. John,
  Ender Konukoglu, Danial Lashkari, José~António Mariz, Raphael Meier,
  Sérgio Pereira, Doina Precup, Stephen~J. Price, Tammy~Riklin Raviv, Syed
  M.~S. Reza, Michael Ryan, Duygu Sarikaya, Lawrence Schwartz, Hoo-Chang Shin,
  Jamie Shotton, Carlos~A. Silva, Nuno Sousa, Nagesh~K. Subbanna, Gabor
  Szekely, Thomas~J. Taylor, Owen~M. Thomas, Nicholas~J. Tustison, Gozde Unal,
  Flor Vasseur, Max Wintermark, Dong~Hye Ye, Liang Zhao, Binsheng Zhao, Darko
  Zikic, Marcel Prastawa, Mauricio Reyes, and Koen Van~Leemput.
\newblock The multimodal brain tumor image segmentation benchmark (brats).
\newblock {\em IEEE Transactions on Medical Imaging}, 34(10):1993--2024, 2015.

\bibitem{milletari2016v}
Fausto Milletari, Nassir Navab, and Seyed-Ahmad Ahmadi.
\newblock V-net: Fully convolutional neural networks for volumetric medical
  image segmentation.
\newblock In {\em Fourth International Conference on 3D Vision (3DV)}, 2016.

\bibitem{peng2017large}
Chao Peng, Xiangyu Zhang, Gang Yu, Guiming Luo, and Jian Sun.
\newblock Large kernel matters--improve semantic segmentation by global
  convolutional network.
\newblock In {\em Proceedings of the IEEE/CVF Conference on Computer Vision and
  Pattern Recognition}, 2017.

\bibitem{monai}
Project-MONAI.
\newblock Medical open network for ai.
\newblock \url{https://github.com/Project-MONAI/MONAI}, 2020.

\bibitem{UNet}
Olaf Ronneberger, Philipp Fischer, and Thomas Brox.
\newblock U-net: Convolutional networks for biomedical image segmentation.
\newblock In {\em International Conference on Medical Image Computing and
  Computer-Assisted Intervention}, 2015.

\bibitem{roth2017hierarchical}
Holger~R Roth, Hirohisa Oda, Yuichiro Hayashi, Masahiro Oda, Natsuki Shimizu,
  Michitaka Fujiwara, Kazunari Misawa, and Kensaku Mori.
\newblock Hierarchical 3d fully convolutional networks for multi-organ
  segmentation.
\newblock {\em arXiv preprint arXiv:1704.06382}, 2017.

\bibitem{MSD}
Amber~L. Simpson, Michela Antonelli, Spyridon Bakas, Michel Bilello, Keyvan
  Farahani, Bram van Ginneken, Annette Kopp-Schneider, Bennett~A. Landman,
  Geert Litjens, Bjoern Menze, Olaf Ronneberger, Ronald~M. Summers, Patrick
  Bilic, Patrick~F. Christ, Richard K.~G. Do, Marc Gollub, Jennifer
  Golia-Pernicka, Stephan~H. Heckers, William~R. Jarnagin, Maureen~K. McHugo,
  Sandy Napel, Eugene Vorontsov, Lena Maier-Hein, and M.~Jorge Cardoso.
\newblock A large annotated medical image dataset for the development and
  evaluation of segmentation algorithms, 2019.

\bibitem{valanarasu2021medical}
Jeya Maria~Jose Valanarasu, Poojan Oza, Ilker Hacihaliloglu, and Vishal~M
  Patel.
\newblock Medical transformer: Gated axial-attention for medical image
  segmentation.
\newblock In {\em International Conference on Medical Image Computing and
  Computer-Assisted Intervention}, 2021.

\bibitem{LinFormer}
Sinong Wang, Belinda~Z Li, Madian Khabsa, Han Fang, and Hao Ma.
\newblock Linformer: Self-attention with linear complexity.
\newblock {\em arXiv preprint arXiv:2006.04768}, 2020.

\bibitem{TransBTS}
Wenxuan Wang, Chen Chen, Meng Ding, Hong Yu, Sen Zha, and Jiangyun Li.
\newblock Transbts: Multimodal brain tumor segmentation using transformer.
\newblock In {\em International Conference on Medical Image Computing and
  Computer-Assisted Intervention}, 2021.

\bibitem{TransFuse}
Yundong Zhang, Huiye Liu, and Qiang Hu.
\newblock Transfuse: Fusing transformers and cnns for medical image
  segmentation.
\newblock In {\em International Conference on Medical Image Computing and
  Computer-Assisted Intervention}, 2021.

\bibitem{zhao2017pyramid}
Hengshuang Zhao, Jianping Shi, Xiaojuan Qi, Xiaogang Wang, and Jiaya Jia.
\newblock Pyramid scene parsing network.
\newblock In {\em Proceedings of the IEEE/CVF Conference on Computer Vision and
  Pattern Recognition}, 2017.

\bibitem{nnFormer}
Hong-Yu Zhou, Jiansen Guo, Yinghao Zhang, Lequan Yu, Liansheng Wang, and Yizhou
  Yu.
\newblock nnformer: Interleaved transformer for volumetric segmentation.
\newblock {\em arXiv preprint arXiv:2109.03201}, 2021.

\bibitem{zhou2018unet++}
Zongwei Zhou, Md~Mahfuzur Rahman~Siddiquee, Nima Tajbakhsh, and Jianming Liang.
\newblock Unet++: A nested u-net architecture for medical image segmentation.
\newblock In {\em Deep learning in medical image analysis and multimodal
  learning for clinical decision support}. 2018.

\bibitem{zhu2017deeply}
Qikui Zhu, Bo Du, Baris Turkbey, Peter~L Choyke, and Pingkun Yan.
\newblock Deeply-supervised cnn for prostate segmentation.
\newblock In {\em International Joint Conference on Neural Networks}, 2017.

\end{thebibliography}
}

\clearpage
\newpage

\appendix
\begin{center}
\textbf{\Large Supplemental Material}
\end{center}
In this section, we provide additional details regarding:
\begin{itemize}\setlength{\itemsep}{-0.5em}
    \item Implementation Details (Appendix~\ref{additional_impl})
    \item Qualitative Results (Appendix~\ref{additional_qual})
    \item Ablations (Appendix~\ref{additional_ablations})
    \item Discussion (Appendix~\ref{discussion})
\end{itemize}

\section{Additional Implementation Details}
\label{additional_impl}
\noindent
\textbf{Overall Architecture}: As presented in Fig.~\ref{fig:Architecture} and described in Sec.~\ref{sec:Overall Architecture}, our architecture consists of a hierarchical encoder-decoder structure. The encoder has four stages in which the number of channels at stages $[C_1,C_2,C_3,C_4]$ are $[32, 64, 128, 256]$ and each stage has three EPA blocks with the number of heads set to four.

Similarly, the decoder has four stages, each consisting of upsampling using deconvolution followed by three EPA blocks. The deconvolutional layers increase the resolution of the feature maps by a factor of two. However, we use a $ 3\times 3 \times 3$ convolutional block at the last stage to compensate the heavy self-attention computation as the spatial size at this stage will be significantly larger (\ie $[128, 128, 64, 16]$ in case of Synapse dataset). The output of the last decoder stage is fused with convolutional features to recover the spatial information and enhance the feature representation. The outputs are then fed into a $3 \times 3 \times 3$ and $1 \times 1 \times 1$ convolutional layers to generate voxel-wise mask predictions.

\noindent
\textbf{Training Details:}
For the Synapse dataset, all the models are trained for 1K epochs with inputs of size 128$\times$128$\times$64. For BTCV, we follow the same training recipe as in~\cite{UNETR} and train all the models at 96$\times$96$\times$96 resolution for 5K epochs. For ACDC, Decathlon-Lung, and BRaTs, we train all the models at 160$\times$160$\times$16, 192$\times$192$\times$34, and 128$\times$128$\times$128 resolutions, respectively. All other training hyper-parameters are same as in~\cite{nnFormer}. Further, we add learnable positional encoding to the input of each EPA block. Our code and pretrained models will be made publicly available to reproduce our results.
\section{Additional Qualitative Results}
\label{additional_qual}
In this section, we provide additional qualitative comparisons for Synapse and ACDC datasets between UNETR++ and the state-of-the-art methods. Moreover, we provide a detailed comparison between UNETR++ and the baseline for Synapse, ACDC, and Dechatlon-Lungs datasets.

\subsection{Synapse Dataset}
\begin{figure*}[!ht]
\centering
{\includegraphics[width=0.90\textwidth]{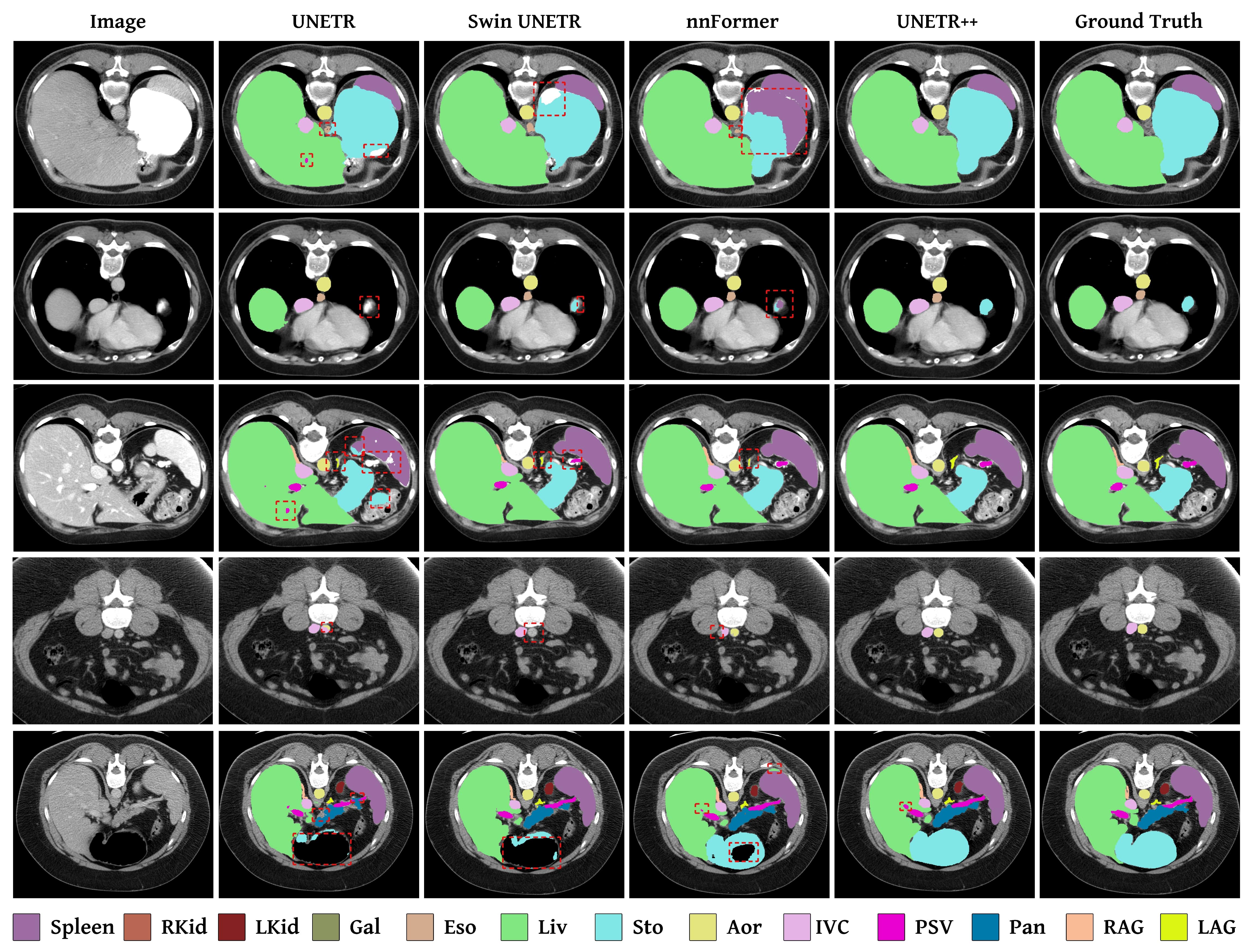}}
\caption{Additional qualitative comparison on Synapse dataset. We compare our UNETR++ with existing methods: UNETR, Swin UNETR, and nnFormer. It is noticeable that the existing methods struggle to correctly segment different organs (marked in red dashed box). Our UNETR++ achieves promising segmentation performance by accurately segmenting the organs. Best viewed zoomed in.}
\label{Fig:Supplementary_Synapse}
\vspace{-0.5cm}
\end{figure*}

Fig.~\ref{Fig:Supplementary_Synapse} shows qualitative comparisons for different cases between UNETR++ and existing approaches on Synapse dataset. The inaccurate predictions are marked with red dashed boxes. In the first and second rows, UNETR++ differentiates the \textit{stomach} tissues at different sizes and \textit{spleen} successfully, while nnFormer struggles to differentiate between \textit{spleen} and \textit{stomach}, UNETR and Swin UNETR struggle to differentiate between \textit{stomach} and the background, demonstrating that UNETR++ provides better segmentation predictions at different scales. In the third row, UNETR++ accurately segments all the organs, while the other existing methods under-segments \textit{left adrenal gland} or \textit{spleen}, and over-segments \textit{stomach} in the case of UNETR. As illustrated in the fourth row, UNETR++ well delineates the boundaries of the \textit{inferior vena cava}, compared to all existing methods which struggle, and confuse it with the background. In the last row, the existing methods under-segments the \textit{stomach} in addition to confusing \textit{pancreas} and \textit{portal and splenic veins} in the case of UNETR. As illustrated, UNETR++ has a better delineation of the boundaries of different organs without under/over-segmenting, thus suggesting that UNETR++ encodes enriched inter-dependent spatial and channel features within the proposed EPA block.

\subsection{ACDC Dataset}
\begin{figure*}[!hb]
\vspace{-0.5cm}
\centering
{\includegraphics[width=0.90\textwidth]{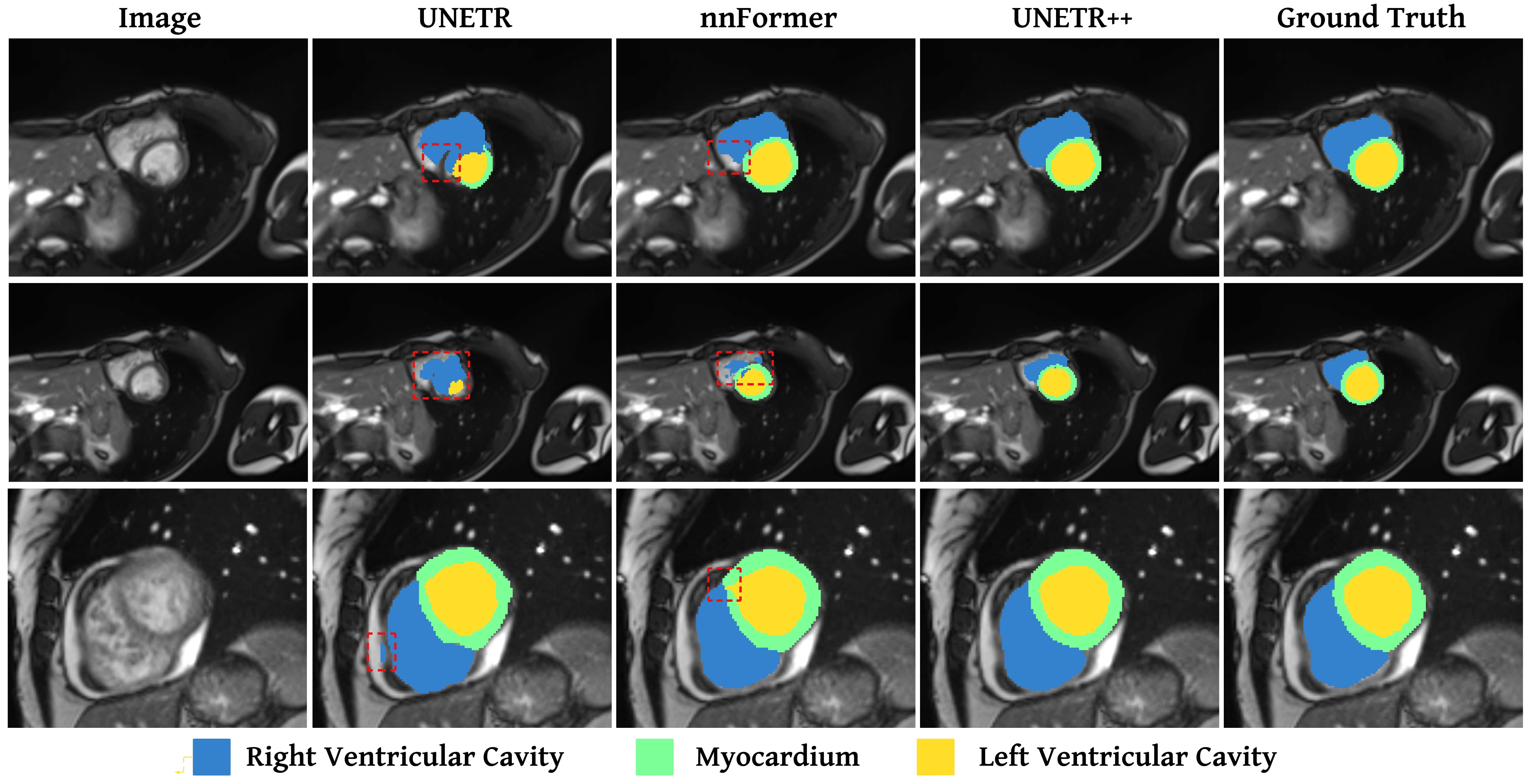}}
\caption{Qualitative comparison on the ACDC dataset. We compare our UNETR++ with existing methods: UNETR and nnFormer. It is noticeable that the existing methods struggle to correctly segment different organs (marked in red dashed box). Our UNETR++ achieves favorable segmentation performance by accurately segmenting the organs. Best viewed zoomed in.}
\label{Fig:Supplementary_ACDC}
\vspace{-0.5cm}
\end{figure*}

Fig.~\ref{Fig:Supplementary_ACDC} shows qualitative comparisons for different cases between UNETR++ and existing approaches, nnFormer and UNETR on the ACDC dataset. The inaccurate predictions are marked with red dashed boxes. In the first row, UNETR and nnFormer under-segments the \textit{right ventricular (RV) cavity}, while our UNETR++ accurately segments all three categories. In the second row, we present a difficult sample where the sizes of all three heart segments are comparatively smaller. In this case, both UNETR and nnFormer under-segments and struggles to delineate between the segments, while UNETR++ gives a better segmentation. In the last row, we present a more simpler sample. However, the existing methods over-segments the \textit{RV cavity} and the \textit{myocardium} in this case, while UNETR++ provides better delineation and provides a segmentation very close to the ground truth. Similar to the observation from Synapse, these qualitative examples shows that, UNETR++ achieves delineation for the three heart segments without under-segmenting or over-segmenting, thus suggesting the importance of its inter-dependent spatial and channel features encoded in the proposed EPA block.

\begin{figure*}[!ht]
\centering
{\includegraphics[width=1\textwidth]{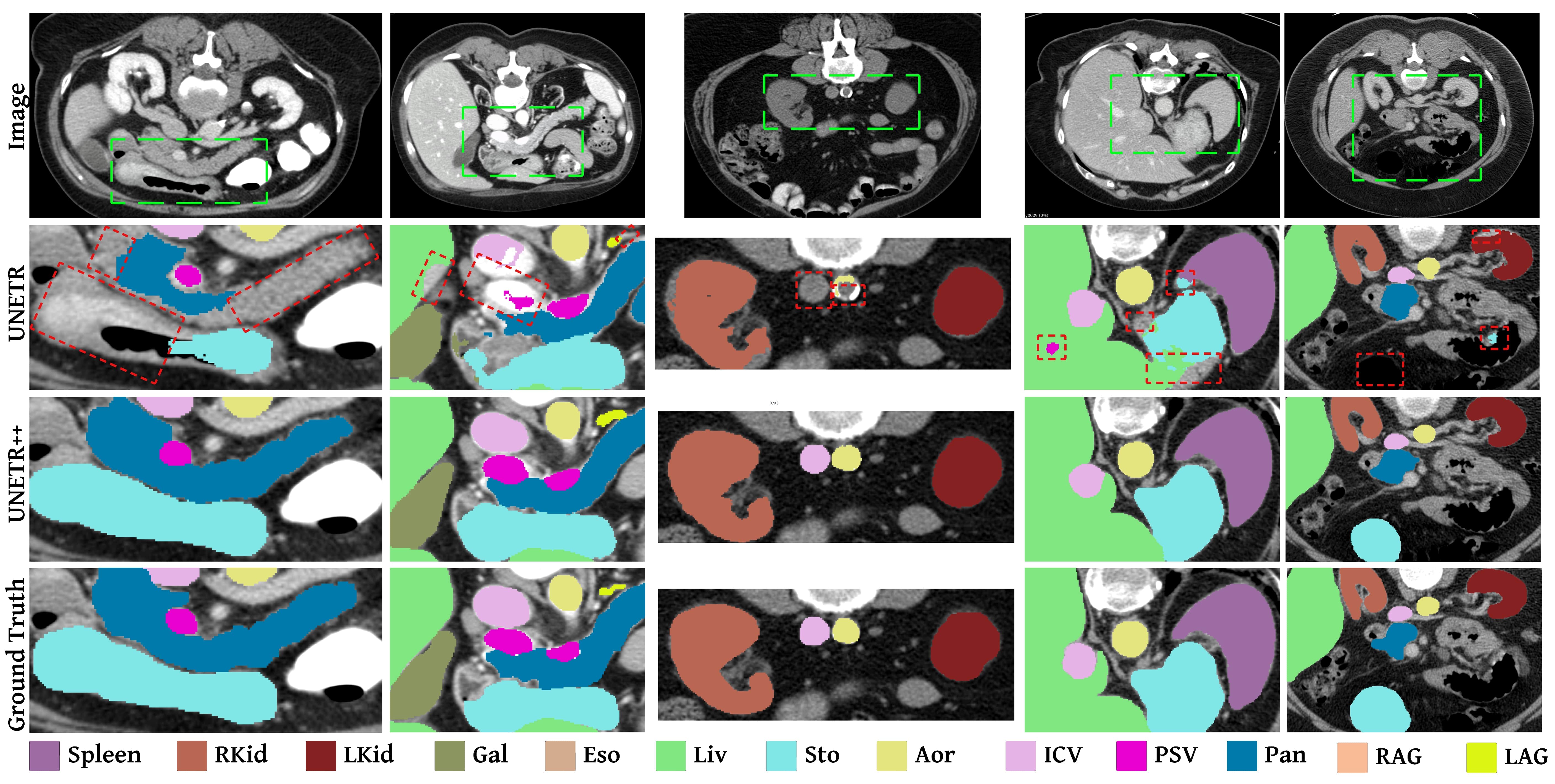}}\vspace{1em}
\caption{Additional qualitative comparison between UNETR++ and the baseline UNETR. The baseline struggle to correctly segment different organs (marked in red dashed box). We enlarge multiple organs (marked with green dashed boxes in the first row) from several cases. Our UNETR++ achieves promising segmentation performance by accurately segmenting the organs. Best viewed zoomed in.}
\label{Fig:Supplementary_Synapse_Baseline}
\vspace{-0.7cm}
\end{figure*}

\begin{figure*}[!b]
\centering
{\includegraphics[width=0.98\textwidth]{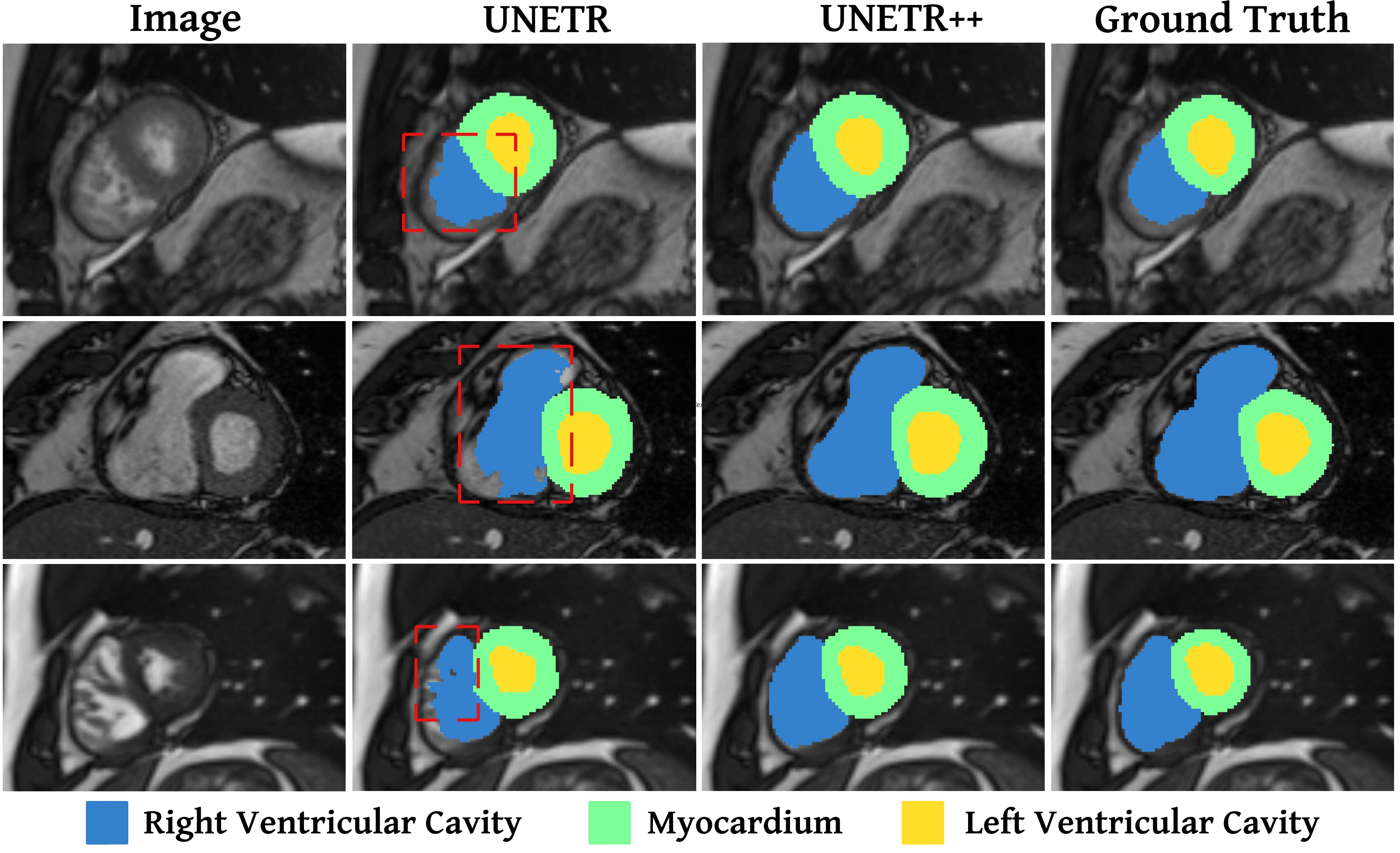}}\vspace{1em}
\caption{Additional qualitative comparison between UNETR++ and the baseline UNETR. The baseline struggle to correctly segment different heart regions (marked in red dashed box). We enlarge the regions from several cases. Our UNETR++ achieves promising segmentation performance by accurately segmenting all regions. Best viewed zoomed in.}
\label{Fig:Supplementary_ACDC_Baseline}
\vspace{-0.5cm}
\end{figure*}

\subsection{Detailed qualitative comparison between UNETR++ and the baseline}

\begin{figure}[t]
  \centering
    \includegraphics[width=1.0\linewidth]{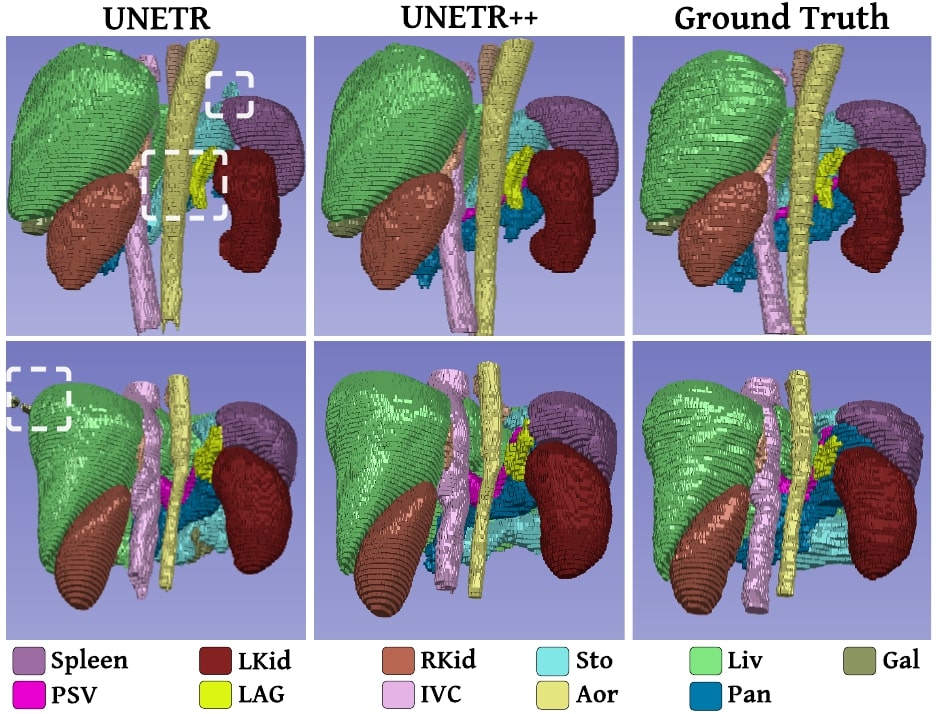}
    \vspace{-0.2in} 
    \caption{Qualitative comparison between the baseline UNETR~\cite{UNETR} and our UNETR++ on Synapse dataset. Each inaccurate segmented region is marked with a white dashed box. UNETR++ better segments the organs as compared to the baseline.}
    \label{fig:Intro_supplemental}
    \vspace{-0.7cm}
\end{figure}

Fig.~\ref{Fig:Supplementary_Synapse_Baseline} shows a qualitative comparison between UNETR++ and the baseline UNETR on Synapse dataset. We present visualizations of enlarged views of different organs (marked with green dashed boxes in the first row) from several cases for better analysis. In the first column, UNETR++ delineates the outline of \textit{pancreas} well, while the baseline notably struggles in segmenting the \textit{pancreas} and under-segments the \textit{stomach}. In the second column, the baseline under-segments \textit{the inferior vena cava} and \textit{portal and splenic veins}, while UNETR++ segments these organs precisely. In the third column, the baseline struggles to segment \textit{the inferior vena cava} and \textit{aorta}. In the last two columns, The baseline under-segments \textit{stomach} and struggles in delineating the boundaries of \textit{spleen} and \textit{left kidney}. On the other hand, UNETR++ achieves improved performance and accurately segments all these organs with better delineation. We further show the 3D rendered segmentation results of UNETR++ in comparison to UNETR in Fig.~\ref{fig:Intro_supplemental}.

\begin{figure}[t]
  \centering
    \includegraphics[width=1.0\linewidth]{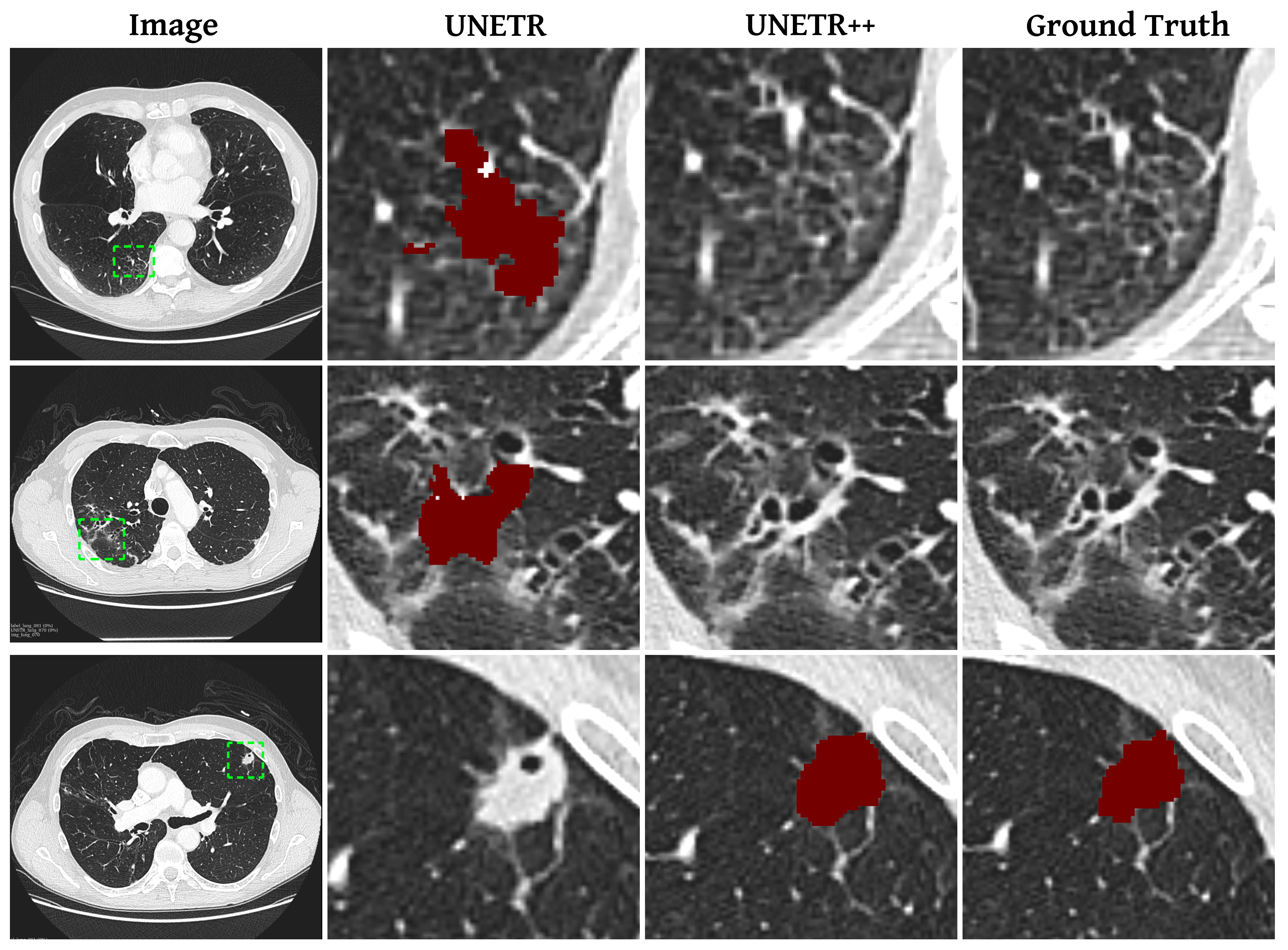}
    \vspace{-0.2in} 
    \caption{Qualitative comparison between the baseline UNETR~\cite{UNETR} and our UNETR++ on Decathlon-Lung dataset. The enlarged area is marked with a green box. UNETR++ has better segmentation and less \textit{false positives} for segmenting the tumors as compared to the baseline. Best viewed zoomed in. }
    \label{Fig:Supplementary_LUNGS_Baseline}
    \vspace{-0.3cm}
\end{figure}

We show in Fig.~\ref{Fig:Supplementary_ACDC_Baseline} qualitative comparison between our UNETR++ and the baseline on the ACDC dataset. In all three rows, UNETR suffers from under-segmentation and struggles in delineating the boundaries of the \textit{right ventricular (RV) cavity}, while UNETR++ segments all three segments more precisely. In addition, we show in Fig.~\ref{Fig:Supplementary_LUNGS_Baseline} another baseline comparison on the Decathalon-Lung dataset. In the first two rows, UNETR++ has less \textit{false positives}, while in the third row, UNETR under-segments the whole tumor and UNETR++ segments it correctly.

\section{Additional Ablations}
\label{additional_ablations}
To investigate the scalability of UNETR++, we designed an experiment with feature maps of size $[64, 128, 256, 512]$ instead of $[32, 64, 128, 256]$ on the BTCV dataset. Although the number of parameters with this change increased to $94.24$M and the FLOPs increased to $117$G, the average dice similarity coefficient (DSC) is improved from $83.28\%$ to $84.27\%$, which proves the scalability of UNETR++ without using any ensemble, pre-training or additional custom data.

To validate the effectiveness of our EPA block,  we conduct experiments on Synapse to compare our EPA module with other attention methods. \textbf{(i)} We integrate the gated attention (GA) from the attention-gated U-Net method within nnUNET (referred to col. 3 in Tab.~\ref{rebuttal:table_2}). \textbf{(ii)} We replace our EPA module in UNETR++, over the proposed hierarchical approach, with GA (col. 4 in Tab.~\ref{rebuttal:table_2} and with squeeze-and-excitation (SE) (col. 5 in Tab. \ref{rebuttal:table_2}). Our UNETR++ achieves superior results compared to other attention methods. 
\begin{table}[t]
\resizebox{\linewidth}{!}{%
\begin{tabular}{cccccc}
\toprule

\multirow{2}{*}{Model} & \multirow{2}{*}{nnUNet} & {Attention} & {EPA replaced } & {EPA replaced } & \multirow{2}{*}\textbf{{\textbf{UNETR++}}}\\
& & nnUNet & w/ GA & w/ SE & \\
\midrule
\midrule
DSC & 84.2 & 85.0 & {85.3} & {85.5} & \textbf{{87.2}}\\
\bottomrule
\end{tabular}}
\vspace{-1em}
\caption{Comparison with other attention methods on Synapse.}
\label{rebuttal:table_2}
\vspace{-1.9em}
\end{table}

\section{Discussion}
\label{discussion}
In this paper, we present a hierarchical approach, named UNETR++, that achieves promising segmentation results on five datasets (Synapse, ACDC, BTCV, BRaTs, and Decathlon-Lung) while significantly reducing the model complexity and the memory consumption, and improving the inference speed compared to existing methods. The proposed efficient paired attention (EPA) block encodes enriched inter-dependent spatial and channel features by using spatial and channel attention. To observe potential limitations of UNETR++, we analyze different outlier cases of Synapse. Although our predictions are better than the existing methods and more similar to the ground truth, we find that there are a few cases where our model, as well as the existing methods, struggle to segment certain organs. When the geometric shape of the organs in a few slices is abnormal (delineated by thin borders), our model and the existing models struggle to segment them accurately. The reason might be the limited availability of training samples with such abnormal shapes compared to the normal samples. We are planning to solve this problem by applying geometric data augmentation techniques at the pre-processing stage.

\end{document}